\documentclass[12pt,a4paper]{article}

\usepackage[margin=1in]{geometry}
\usepackage[T1]{fontenc}
\usepackage{setspace}
\usepackage{mathtools,amssymb,amsthm,bm}
\usepackage{mathrsfs}
\usepackage{graphicx}
\usepackage{booktabs}
\usepackage{multirow}
\usepackage{longtable}
\usepackage{pdflscape}
\usepackage{placeins}
\makeatletter
\renewcommand{\subsection}{\@startsection{subsection}{2}{\z@}%
  {-2.5ex plus -1ex minus -.2ex}{.8ex plus .2ex}%
  {\normalfont\large\bfseries}}
\makeatother
\usepackage{etoolbox}
\makeatletter
\patchcmd{\l@section}{1.5em}{2.1em}{}{}
\makeatother
\usepackage{xcolor}
\usepackage{algorithm,algpseudocode}

\usepackage{enumerate}
\usepackage[authoryear,round]{natbib}
\usepackage[hidelinks]{hyperref}

\theoremstyle{plain}
\newtheorem{theorem}{Theorem}
\newtheorem{proposition}{Proposition}
\newtheorem{corollary}{Corollary}
\newtheorem{lemma}{Lemma}
\theoremstyle{definition}
\newtheorem{definition}{Definition}
\newtheorem{assumption}{Assumption}
\theoremstyle{remark}
\newtheorem{remark}{Remark}

\newcommand{\R}{\mathbb{R}}
\newcommand{\E}{\mathbb{E}}
\newcommand{\Prob}{\mathbb{P}}
\newcommand{\Cov}{\mathrm{Cov}}
\newcommand{\tr}{\mathrm{tr}}
\newcommand{\diag}{\mathrm{diag}}
\newcommand{\op}{\mathrm{op}}
\newcommand{\Exp}{\mathrm{Exp}}
\newcommand{\Log}{\mathrm{Log}}
\newcommand{\Sym}{\mathrm{Sym}}
\newcommand{\Hess}{\mathrm{Hess}}
\newcommand{\eps}{\varepsilon}
\newcommand{\del}{\delta}
\newcommand{\ip}[2]{\left\langle #1,#2\right\rangle}
\newcommand{\norm}[1]{\left\lVert #1\right\rVert}
\newcommand{\ot}{\otimes}
\newcommand{\F}{\mathcal{F}}
\newcommand{\cY}{\mathcal{Y}}

\newcommand{\SPD}{\mathbb{P}}
\newcommand{\sph}{\mathbb{S}}
\newcommand{\Normal}{\mathsf{N}}
\newcommand{\HH}{\mathbb{H}}
\newcommand{\PhiN}{\Phi}

\title{Locally Private Inference for Riemannian Stochastic Optimization}
\author{Xiaotian Chang\thanks{School of Physical and Mathematical Sciences, Nanyang Technological University},
Yangdi Jiang\footnotemark[1], and
Qirui Hu\thanks{\begin{tabular}[t]{@{}l@{}}
School of Statistics and Data Science, Shanghai University of Finance and Economics.\\
Department of Mathematics, Ruhr-Universit\"at Bochum\\
Corresponding author. Email: \href{mailto:huqirui@mail.shufe.edu.cn}{huqirui@mail.shufe.edu.cn}
\end{tabular}}}
\date{}

\begin{document}
\maketitle
\begin{abstract}
We develop inference for manifold-valued population minimizers when each
observation belongs to a different participant and only locally private
messages reach the analyst.  The method releases randomized tangent gradients
and combines them through Riemannian stochastic approximation and
Polyak--Ruppert averaging.  Directly inserting a private data surrogate into a
nonlinear loss can shift its population target, whereas conditional centring
of the released gradient preserves the first-order equation.  We introduce
symmetric-pair regression (SPR) to estimate the asymptotic variance from the
same private messages used for point estimation, without holding out
participants or requesting a second release.  We prove the central
limit theorem and consistency of the fully transcript-based sandwich
covariance and intrinsic Wald region under local differential privacy.
Simulations across various  statistical problems and  manifolds
support the predicted decrease in estimation error and near-nominal coverage
under moderate privacy.
An application to NHANES anthropometric data illustrates private estimation
of a leading body-size direction and its uncertainty.
\end{abstract}


\onehalfspacing
\raggedbottom
\setlength{\abovedisplayskip}{8pt plus 2pt minus 2pt}
\setlength{\belowdisplayskip}{8pt plus 2pt minus 2pt}
\setlength{\abovedisplayshortskip}{4pt plus 2pt minus 1pt}
\setlength{\belowdisplayshortskip}{6pt plus 2pt minus 2pt}
\setlength{\jot}{4pt}

\clearpage
\section{Introduction}\label{sec:intro}

Geometric data and constrained parameters play an important role in modern
statistics, while the sensitivity of medical and health measurements makes
privacy a central concern.  Diffusion-tensor imaging represents local tissue
structure by positive-definite matrices, and population summaries must
respect this matrix geometry \citep{Arsigny2006}.
Figure~\ref{fig:intro-dti-cone} illustrates this setting and the use of
locally private scores for geometric estimation and inference.
In health surveys, a principal direction of body measurements describes
joint variation in height, weight, and waist circumference; the parameter is
a direction on a sphere.  Such applications motivate geometric methods:
intrinsic Fr\'echet means minimise expected squared geodesic distance;
directional locations and principal components are constrained targets on
spheres or quotient manifolds; covariance matrices admit affine-invariant and
log-Euclidean formulations on the positive-definite cone
\citep{BhattacharyaPatrangenaru2005,Pennec2006,Afsari2011,Edelman1998,JolliffeCadima2016,Arsigny2007}.
These examples share a population objective
\[
 F(x)=\E\{\ell(x,Y)\},\qquad x\in M,
\]
whose locally unique minimiser \(x_\star\) lies on a Riemannian manifold
\(M\).

% The argument is retained for compatibility with both manuscript versions.
\newcommand{\IntroDTIConeCaption}[1]{%
Diffusion tensors and locally private geometric inference.
Left and centre: a Stanford HARDI mean-$b_0$ image and fitted diffusion
tensors from one participant (Rokem et al., 2013; PDDL~1.0).
The LDP arrow represents participants retaining their tensors and releasing
randomised tangent scores for estimation and inference.
Right: synthetic tensors, their affine-invariant mean, and a schematic
uncertainty region in an illustrative $\mathrm{SPD}(2)$ cone.}

\begin{figure}[!htbp]
  \centering
  \includegraphics[width=\textwidth]{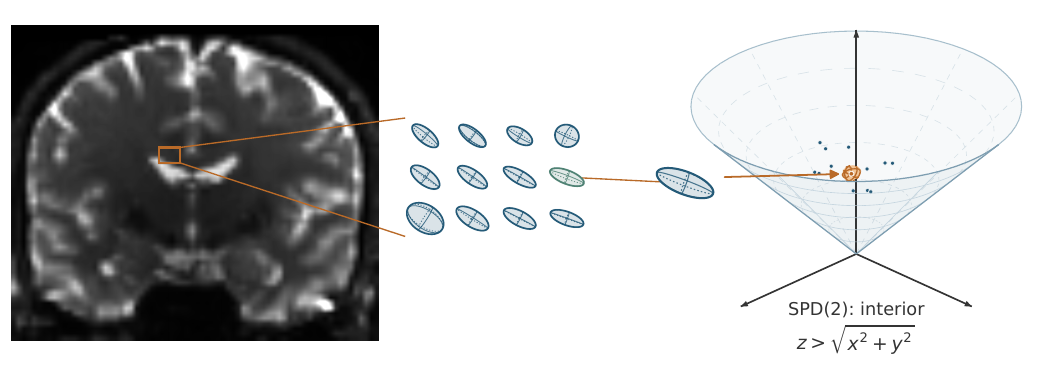}
  \caption{\IntroDTIConeCaption{sec:dti-construction}}
  \label{fig:intro-dti-cone}
\end{figure}

Riemannian stochastic gradient descent (RSGD) is a natural one-pass estimator when
observations arrive sequentially.  Retractions and parallel transport respect
the geometry of the parameter space \citep{Absil2008,Boumal2023}, and
Polyak--Ruppert averaging recovers root-\(n\) statistical efficiency under
local smoothness and stability conditions
\citep{Bonnabel2013,Tripuraneni2018}.  Local differential privacy adds a
distinct constraint: the analyst cannot retain or revisit an observation, and
successive contributions live in tangent spaces attached to changing public
queries.

Replacing each observation by a private surrogate and then applying the same
loss can introduce bias by shifting the population target.  This is not peculiar to
Riemannian optimization: the same issue arises for Euclidean estimators and
stochastic-gradient methods whenever the score is nonlinear in the data.
For regression with perturbed data, \citet{JiangMeasurement2024} connect
this bias to measurement error and construct corrections that account for
the privacy mechanism.
Manifold geometry introduces further nonlinearities, but the basic
compatibility requirement is that the original target must still solve the
privatised first-order equation.  Section~\ref{sec:target-shift} gives examples
where this requirement holds and explains why it often fails. Our inference procedure privatises stochastic gradients instead.  At a public
query, one participant computes a bounded or clipped tangent score and
releases a single randomised vector.  Conditional unbiasedness preserves the
population estimating equation.  The main difficulty is feasible
studentisation: the analyst must recover both the score covariance and the
full Hessian from the same stream of private messages.  The gradients alone
do not guarantee enough variation in every direction to estimate the Hessian,
and the analyst has no access to the observations needed to compute second
derivatives.  Resolving this difficulty is essential for turning
a private point estimate into a confidence region whose calibration can be
computed from the available data.

\begin{figure}[!htbp]
\centering
\includegraphics[width=\textwidth]{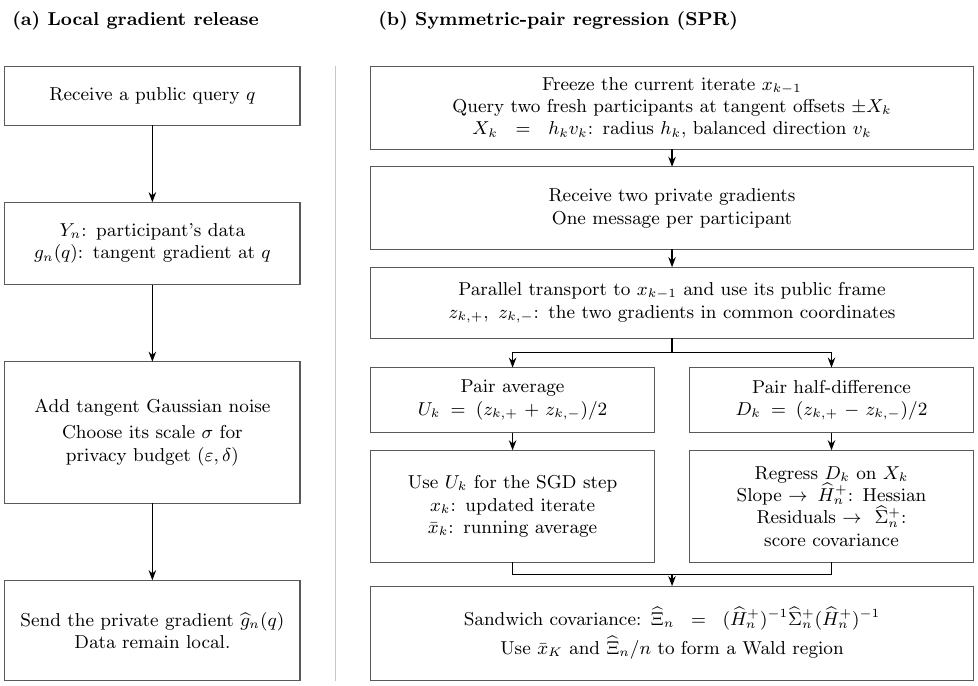}
\caption{Private gradient release and symmetric-pair regression (SPR).
Left: each participant releases one gradient with locally generated Gaussian noise.
Right: messages at opposite offsets $\pm X_k$ are brought to common coordinates $z_{k,+}$ and $z_{k,-}$.
Their average $U_k$ updates the point estimate; regression of their half-difference $D_k$ on $X_k$ estimates the Hessian and score covariance.
The superscript $+$ denotes the positive-definite estimates used in the sandwich covariance.}
\label{fig:spr-overview}
\end{figure}

We address this problem with symmetric-pair regression (SPR), a variance
estimator based on balanced symmetric queries.  The server freezes an iterate
and queries two fresh participants at equal and opposite tangent offsets.  After transport to the frozen base, the message
average drives the optimization update.  The message difference is a noisy
central difference of the population gradient and therefore identifies the
Hessian.  Cycling the offset direction through a public tight frame makes the
regression Gram matrix a scalar multiple of the identity.  This gives
controlled Hessian information even when the realised optimization path
has little variation in some directions.  Both messages in each pair are
available for the point recursion and the nuisance estimates, so Hessian
estimation does not reduce the sample available for point
estimation or require another release from a participant.
Figure~\ref{fig:spr-overview} summarises how each private message supports both estimation and inference; see Sections \ref{sec:ldp-randomisation} and \ref{sec:end-to-end-algorithm} for details.

Finite differences of noisy gradients and simultaneous-perturbation Hessian
estimation are classical tools in Euclidean stochastic approximation
\citep{Ruppert1985,Spall2000}.  Using these ideas for intrinsic inference from
a single locally private transcript poses a further theoretical challenge:
we must
balance the finite-difference bias against the information available for
Hessian estimation while controlling changing tangent spaces and the
dependence between the point and nuisance estimates.  We establish stability,
an all-user Polyak--Ruppert limit, and consistency of both nuisance estimates
under checkable local conditions.  Together these results yield intrinsic
Wald regions computed entirely from private messages, with sampling and
privacy variation both represented in the estimated covariance.

The work connects intrinsic large-sample theory, Riemannian stochastic
approximation, and local privacy.  Intrinsic-mean inference supplies the local
chart and Hessian structure
\citep{BhattacharyaPatrangenaru2003,BhattacharyaPatrangenaru2005};
Riemannian stochastic approximation supplies the averaged recursion
\citep{Bonnabel2013,Tripuraneni2018}; and local privacy supplies the
message-level randomisation and sensitivity calibration
\citep{KairouzOhViswanath2016,DuchiJordanWainwright2018}.  Private manifold
methods have studied central or output perturbations and empirical
optimization, with recent work on feasible centrally private inference for
Fr\'echet functionals
\citep{Reimherr2021,Soto2022,Utpala2023,Jiang2023,Han2024,JiangChangHu2026,ChangEtAl2026RDP}.
These developments explain how manifold geometry shapes private releases;
our focus is on estimating uncertainty from sequential local releases.

In Euclidean settings, online private inference has been developed for
stochastic-gradient procedures in privacy-preserving settings \citep{Liu2023,XieEtAl2025,CaiHuSunWu2025,CaiHuWu2026Federated}.
%\citep{CaiHuWu2026Integration}.
These works show how the structure of an
estimating problem can make private inference feasible.  Here symmetric
queries identify a full Hessian and score covariance in changing tangent
spaces, giving an intrinsic sandwich estimator from the same messages used
for point estimation.

Numerical experiments across six model classes examine estimation error and
coverage for the complete procedure.  Comparisons with trajectory regression
and half-sample probes show how controlling excitation and retaining all
users affect the accuracy of inference.  An anthropometric
principal-direction analysis illustrates how private tangent messages can
support uncertainty statements about a scientifically interpretable target.

The rest of the paper is organised as follows.
Section~\ref{sec:method} introduces the privacy mechanism, SPR,
and assumptions.  Section~\ref{sec:theory} establishes the
inferential guarantees and their main special cases.
Section~\ref{sec:model-classes} develops the statistical applications,
Section~\ref{sec:sim} reports simulations, and Section~\ref{sec:nhanes}
presents the anthropometric analysis.  Section~\ref{sec:conclusion}
concludes.  Appendices~\ref{app:model-examples}--\ref{app:auxiliary-results} provide
further results and full simulation specifications.  Proofs are given in
the Supplementary Material.

\section{Methodology}\label{sec:method}

\subsection{Geometric and statistical setup}\label{sec:setup}

Let $(M,g)$ be a complete $d$-dimensional Riemannian manifold. For $x\in M$, write $T_xM$ for the tangent space and $\norm{\cdot}_x$ for the norm induced by $g$ on $T_xM$. Let $d(\cdot,\cdot)$ denote geodesic distance.
The exponential map $\Exp_x(v)$ follows the geodesic starting at $x$ with velocity $v$ for unit time; its local inverse is $\Log_x$.
We work in a normal convex neighbourhood, where any two points are joined by a unique minimising geodesic within the neighbourhood.
Parallel transport moves tangent vectors along this geodesic while preserving inner products; we write it as $\Gamma_y^x:T_yM\to T_xM$.

The parameter is the locally unique minimiser
\begin{equation*}
        x_\star\in \operatorname*{argmin}_{x\in K} F(x),
        \qquad F(x):=\E[\ell(x,Y)],
\end{equation*}
of a smooth population objective $F$ on a compact working region $K\subset M$ with $x_\star\in\mathrm{int}(K)$. The observation $Y$ takes values in a measurable space $\cY$. The stochastic gradient at a query point $x$ is
\begin{equation*}
        g_n(x):=\nabla_x\ell(x,Y_n)\in T_xM .
\end{equation*}
The algorithm uses a publicly chosen second-order retraction $R$, a smooth approximation to $\Exp$ that agrees with it through second order at zero.
Its local inverse $R_x^{-1}(y)$ represents a nearby point $y$ as a tangent vector at $x$.
Either $R_x^{-1}$ or $\Log_x$ may be used for the Wald region: at root-$n$ distances from $x$, their difference is $o(n^{-1/2})$.

The procedure below updates once for every two participants.  At each update,
two private gradients are transported to one frozen public base and averaged,
and the resulting tangent vector drives a Riemannian stochastic-approximation
step.  The point estimator is the recursive intrinsic Polyak--Ruppert average.
We use \(x_k\) for the update sequence, \(\bar x_k\) for its average, and
\(\Delta_k=R_{x_\star}^{-1}(x_k)\) for local calculations.  Coordinate
representations are taken in the smooth public frame fixed in
Section~\ref{sec:ldp-randomisation}.

\phantomsection\label{sec:target-shift}
A natural first attempt is to release a private surrogate \(Z\) for each
observation and insert \(Z\) into the original loss.  This works when the
privacy channel preserves, or permits recovery of, the estimating equation.
For example, a mean-preserving vector mechanism preserves the squared-error
target for a bounded Euclidean mean \citep{DuchiJordanWainwright2018}.
For positive-definite matrix data, the log-Euclidean mean is obtained by averaging the matrix logarithms and exponentiating the result.
A mean-preserving private release of bounded matrix logarithms therefore preserves this target.  For categorical data, randomised response followed by
inversion of its known channel recovers the original cell probabilities
\citep{Warner1965,KairouzOhViswanath2016}.

In these examples, the score is affine in the released quantity or the effect of randomisation can be explicitly corrected.  With a general nonlinear loss, a centred perturbation of the data
need not give a centred perturbation of the score, so the minimiser of the
privatised objective can differ from \(x_\star\).  This is a statistical
problem rather than a feature of Riemannian optimization: it arises for
Euclidean \(M\)-estimators and stochastic-gradient methods as well.
Appendix~S.4 characterises the resulting target shift
and gives its small-noise expansion.

We instead randomise the tangent score evaluated at each public query.
Conditional centring then preserves the population first-order equation.
The released scores also contain exactly the local first-order information
used below: pair averages drive the recursion, while pair differences recover
the Hessian and score covariance required for inference.

\subsection{Interactive local privacy and tangent-space randomisation}\label{sec:ldp-randomisation}

The privacy model is one-pass and interactive. At time $n$, the server broadcasts the current public query point $x_{n-1}$. User $n$ observes $(x_{n-1},Y_n)$, computes the stochastic gradient $g_n(x_{n-1})$, applies a local randomiser, and sends back one message. Since each user participates once, there is no within-user composition across time. Future public iterates, covariance estimation, and confidence-set construction are functions of already privatised messages, other users' data through their own privatised messages, and public randomness; they are therefore post-processing for the participant whose privacy is being analysed.

For a query $x$ and datum $y$, write $Q_x(A\mid y)$ for the probability that the randomised message lies in $A$.
This family of probability distributions, measurable in its inputs, is called a Markov kernel.
\begin{definition}[Interactive local differential privacy]\label{def:ldp}
A family of Markov kernels $\{Q_x:x\in K\}$ from $\cY$ to query-dependent measurable output spaces $(\mathcal Z_x,\mathcal A_x)$ is $(\eps,\del)$-locally differentially private (or $(\eps,\del)$-LDP) if, for every fixed public query $x\in K$, every $y,y'\in\cY$, and every $A\in\mathcal A_x$,
\begin{equation*}
        Q_x(A\mid y)\le e^\eps Q_x(A\mid y')+\del .
\end{equation*}
\end{definition}

The transcript $\mathcal T_N$ is the complete public record of queries, released messages, and public randomness from $N$ participants.
All subsequent public computations are measurable functions of this record, and each participant's data enter only through that participant's message.
\begin{proposition}\label{prop:adaptive-transcript}
In this one-message protocol, suppose each conditional message distribution satisfies Definition~\ref{def:ldp} for every realised prior public history.
Then, for every participant $i\le N$, fixed data $y_{-i}$ of the other participants, $y,y'\in\cY$, and measurable transcript event $A$,
\begin{equation}\label{eq:adaptive-transcript-ldp}
        \Prob(\mathcal T_N\in A\mid Y_i=y,Y_{-i}=y_{-i})
        \le e^\eps
        \Prob(\mathcal T_N\in A\mid Y_i=y',Y_{-i}=y_{-i})
        +\del .
\end{equation}
Thus \(\mathcal T_N\) is user-level \((\eps,\del)\)-LDP for each participant.
\end{proposition}

For each public query \(x\), let \(B_x:\R^d\to T_xM\) denote the
public orthonormal frame used for tangent coordinates. Choose an orthonormal basis at one public reference point and parallel transport each basis vector along the geodesic to $x$.
The resulting basis defines $B_x$ smoothly throughout the normal convex working neighbourhood. Euclidean randomisers
act on \(B_x^{-1}g_n(x)\).

Let $\F_{n-1}$ denote the public information available before participant $n$ responds, including the current query and the chosen randomiser.
\begin{proposition}\label{prop:lift}
Suppose $\norm{g_n(x)}_x\le G$ for all $x\in K$ and all possible data values. Let $R_{\eps,\del}$ be a Euclidean randomiser on $\R^d$ such that
\begin{enumerate}[(a)]
\item $R_{\eps,\del}$ is $(\eps,\del)$-LDP on the Euclidean ball $\{v:\norm v\le G\}$;
\item $\E[R_{\eps,\del}(v)]=v$ for every $\norm v\le G$.
\end{enumerate}
Define $
        \widehat g_n(x):=B_xR_{\eps,\del}\{B_x^{-1}g_n(x)\}\in T_xM .
$
Then, for every fixed public $x\in K$, the map $Y_n\mapsto\widehat g_n(x)$ is $(\eps,\del)$-LDP and
$
        \E\left[\widehat g_n(x)\,\middle|\, Y_n,\F_{n-1}\right]=g_n(x).
$
\end{proposition}

The most transparent randomiser is the tangent Gaussian mechanism. Let $\Delta_{\mathrm{sens}}=2G$ be the local sensitivity of the tangent-gradient query at a fixed public point. For arbitrary $\eps>0$, the analytic Gaussian calibration can be used: choose $\sigma$ so that
\begin{equation}\label{eq:analytic-gaussian-calib}
        \PhiN\!\left(\frac{\Delta_{\mathrm{sens}}}{2\sigma}-\frac{\eps\sigma}{\Delta_{\mathrm{sens}}}\right)
        -e^\eps\PhiN\!\left(-\frac{\Delta_{\mathrm{sens}}}{2\sigma}-\frac{\eps\sigma}{\Delta_{\mathrm{sens}}}\right)
        \le \del,
\end{equation}
where $\PhiN$ is the standard normal distribution function \citep{BalleWang2018}. A simpler sufficient, but generally non-optimal, classical scale is $
        \sigma\ge 2G\sqrt{2\log(1.25/\del)}/{\eps},
$
under the usual small-$\eps$ conditions for the classical Gaussian mechanism.

\begin{proposition}\label{prop:tangent-gauss}
Suppose $\norm{g_n(x)}_x\le G$ on $K$. Let $Z_n\sim\Normal(0,I_d)$ be independent of $(Y_n,\F_{n-1})$ and define
\begin{equation}\label{eq:tangent-gauss}
        \widehat g_n(x)=g_n(x)+\sigma B_xZ_n .
\end{equation}
If $\sigma$ satisfies \eqref{eq:analytic-gaussian-calib} with $\Delta_{\mathrm{sens}}=2G$, then $Y_n\mapsto\widehat g_n(x)$ is $(\eps,\del)$-LDP for every fixed $x\in K$. Moreover, $
        \E\left[\widehat g_n(x)\,\middle|\, Y_n,\F_{n-1}\right]=g_n(x)$, $
        \Cov\left\{\widehat g_n(x)-g_n(x)\,\middle|\, Y_n,\F_{n-1}\right\}=\sigma^2 I_{T_xM}$ and $
        \E\norm{\widehat g_n(x)-g_n(x)}_x^4=\sigma^4d(d+2).$
\end{proposition}

\subsection{Symmetric-pair regression and private inference}\label{sec:private-procedure}
\label{sec:end-to-end-algorithm}

The procedure returns a point estimate and a confidence region from
one private gradient per participant.  The leading covariance of the averaged
estimate has the sandwich form \(H^{-1}\Sigma_{\mathrm{tot}}H^{-1}/n\):
\(H\) is the Hessian of \(F\) at \(x_\star\), and
\(\Sigma_{\mathrm{tot}}\) is the limiting covariance of a released gradient there,
both expressed in the public frame.  SPR estimates these two matrices
using the construction in Figure~\ref{fig:spr-overview}.  The average of
two messages updates the point estimate; their difference measures how the
mean gradient changes with the query, and hence estimates the Hessian.
The unexplained variation in those differences estimates the score
covariance.  Algorithm~\ref{alg:spr} collects the steps below.

\noindent\textbf{Paired private queries.}
For clarity, take \(n=2K\); with an odd number of participants, the final
participant may supply one ordinary update.  At pair \(k\), freeze the
current iterate \(x_{k-1}\).  Two fresh participants evaluate gradients at
opposite offsets of radius \(h_k\) in a public unit direction
\(v_k\in\R^d\):
\begin{equation}\label{eq:pair-queries}
 q_{k,+}=R_{x_{k-1}}\{h_kB_{x_{k-1}}v_k\},\qquad
 q_{k,-}=R_{x_{k-1}}\{-h_kB_{x_{k-1}}v_k\}.
\end{equation}
Both queries and both mechanism choices are frozen in the public
history \(\mathcal G_{k-1}\) before either participant responds.
Each fresh participant releases one gradient through the mechanism of
Section~\ref{sec:ldp-randomisation}, using independent fresh randomisation.  Because the gradients belong to different
tangent spaces, the server first transports them to \(x_{k-1}\) and
expresses them in its frame.  It then forms their average \(U_k\) and
half-difference \(D_k\), retaining the known offset \(X_k\):
\begin{align}
 z_{k,+}&=B_{x_{k-1}}^{-1}\Gamma_{q_{k,+}}^{x_{k-1}}
              \widehat g_{2k-1}(q_{k,+}),\notag\\
 z_{k,-}&=B_{x_{k-1}}^{-1}\Gamma_{q_{k,-}}^{x_{k-1}}
              \widehat g_{2k}(q_{k,-}),\label{eq:pair-messages}\\
 U_k&=\frac{z_{k,+}+z_{k,-}}2,\qquad
 D_k=\frac{z_{k,+}-z_{k,-}}2,\qquad X_k=h_kv_k.\notag
\end{align}

\noindent\textbf{Point estimation.}
The symmetric average \(U_k\) approximates the gradient at the current
iterate: the first-order effects of the positive and negative offsets
cancel.  We use it in a Riemannian gradient step and recursively average the
iterates.  To keep the queries and updates in the working region, choose
public compact sets \(K_{\rm s}\subset\operatorname{int}(K_{\rm q})\),
with \(K_{\rm q}\subset K\), for iterates and queries, respectively.
The sets and initialization \(x_0=\bar x_0\in K_{\rm s}\) are fixed
publicly before collecting releases.  Choose a public upper bound on
\(h_k\) so that each entire path
\(R_x(\pm tB_xv_k)\), \(0\le t\le h_k\), lies in \(K_{\rm q}\).
The default safeguard is the fixed metric projection

\[
 \Pi(y)=P_{K_{\rm s}}(y)
 :=\operatorname*{argmin}_{z\in K_{\rm s}}d(y,z)^2.
\]

It is single-valued on a sufficiently small public tubular
neighbourhood of \(K_{\rm s}\).  Choose the public cap \(\delta_0\) so
that \(R_x(u)\) lies in this neighbourhood for every
\(x\in K_{\rm s}\) and \(\norm{u}\le\delta_0\).
With gain \(\gamma_k\), define
\(\widetilde U_k=\mathsf C_{\delta_0/\gamma_k}(U_k)\), where
\(\mathsf C_\tau(u)=u\min\{1,\tau/\|u\|\}\) and
\(\mathsf C_\tau(0)=0\).  Starting from \(x_0=\bar x_0\), compute
{
\begin{align}
 y_k&=R_{x_{k-1}}\{-\gamma_kB_{x_{k-1}}\widetilde U_k\},
       \qquad x_k=\Pi(y_k),\notag\\
 \bar x_k&=\Exp_{\bar x_{k-1}}\!\left\{\frac1k
                         \Log_{\bar x_{k-1}}(x_k)\right\}.
 \label{eq:pair-update}
\end{align}
}
Thus \(x_k\) is the current iterate and \(\bar x_K\) is the reported
point estimate.  The cap acts only on the update; the original differences
\(D_k\) remain available for regression.

\noindent\textbf{Hessian and score covariance estimation.}
At a base \(x=x_{k-1}\) near \(x_\star\), a first-order expansion in the
common tangent frame gives
\[
 \E(D_k\mid\mathcal G_{k-1})
 \approx \tfrac12\bigl[\{B_x^{-1}\nabla F(x)+HX_k\}
                    -\{B_x^{-1}\nabla F(x)-HX_k\}\bigr]=HX_k.
\]
The base gradient cancels, so regressing \(D_k\) on \(X_k=h_kv_k\)
estimates \(H\) from gradient messages alone.
For the assumptions and analysis, write the transported mean field as

\begin{equation}\label{eq:Psi-field}
 \Psi_x(w)=B_x^{-1}\Gamma_{R_x(B_xw)}^x
              \nabla F\{R_x(B_xw)\},\qquad w\in\R^d.
\end{equation}

For \(i\in\{2k-1,2k\}\), also let

\[
 C_i(x,q)=\Cov\!\left(B_x^{-1}\Gamma_q^x\widehat g_i(q)
                         \mid\mathcal G_{k-1}\right).
\]

Here the conditional distribution of $\widehat g_i(q)$ averages over the fresh participant's data and randomisation, holding the public history fixed.
We take this distribution and its means and covariances to be jointly measurable in the history and query, and in the base point after transport.

The public directions repeat a prescribed finite cycle
\(\{v_1,\ldots,v_R\}\) satisfying

\begin{equation}\label{eq:tight-frame-design}
 \sum_{r=1}^Rv_r=0,\qquad
 \frac1R\sum_{r=1}^Rv_rv_r^{\mathsf T}=\frac1dI_d,\qquad
 \norm{v_r}=1.
\end{equation}

A simple choice is \(e_1,-e_1,\ldots,e_d,-e_d\), where
\(e_j\) is the \(j\)th coordinate vector.  For \(v_k=e_j\), the mean of \(D_k/h_k\) approximates \(He_j\),
the \(j\)th column of \(H\).  Each cycle entry uses two fresh participants
at both offsets, so \(e_j\) and \(-e_j\) specify separate pairs.
The deterministic radii are constant within each cycle, giving \(\sum X_k=0\) and an
isotropic design matrix, providing equal regression information in every
tangent direction.  The radii may decrease between cycles; radii and
directions are prescribed before observing the messages.

For the regression, use a window fixed by the public clock, omitting an initial fraction of pairs that tends to zero and retaining only complete cycles.  Denote their index set by
\(\mathcal I_n\), its size by \(L_n\), and total squared radius by
\(s_n=\sum_{k\in\mathcal I_n}h_k^2\).  All pairs still contribute to the
point estimate.  The fitted intercept \(\bar D_n\) and slope
\(\widehat B_n\) solve
\[
 \left(\bar D_n,\widehat B_n\right)
 =\underset{a\in\R^d,\,B\in\R^{d\times d}}{\operatorname{arg\,min}}
 \left\{\sum_{k\in\mathcal I_n}\norm{D_k-a-BX_k}^2
                 +\lambda_n\norm{B}_{\mathrm F}^2\right\}.
\]
Here \(\norm{\cdot}_{\mathrm F}\) is the Frobenius norm and
\(\lambda_n\ge0\) penalises only the slope, leaving the intercept
unpenalised.  Since \(\sum_{k\in\mathcal I_n}X_k=0\), the closed forms are
\begin{align}
 \bar D_n&=L_n^{-1}\sum_{k\in\mathcal I_n}D_k,\qquad
 Q_n=\sum_{k\in\mathcal I_n}X_kX_k^{\mathsf T},\notag\\
 A_n&=\sum_{k\in\mathcal I_n}(D_k-\bar D_n)X_k^{\mathsf T},\qquad
 \widehat B_n=A_n(Q_n+\lambda_nI_d)^{-1}.
 \label{eq:pair-regression}
\end{align}
Symmetrising the slope gives the Hessian estimate
\(\widehat H_n^{\mathrm{raw}}=\left(\widehat B_n+\widehat B_n^{\mathsf T}\right)/2\).
Next compute residuals \(e_k=D_k-\bar D_n-\widehat B_nX_k\) and set
\begin{equation}
 \widehat\Sigma_n^{\mathrm{raw}}
 =\frac{2}{L_n}\sum_{k\in\mathcal I_n}e_ke_k^{\mathsf T}.
 \label{eq:pair-covariance}
\end{equation}
The factor \(2\) converts the covariance of a half-difference back to
that of one released score: the independent noise terms in two messages,
each with limiting covariance \(\Sigma_{\mathrm{tot}}\), give a
half-difference with covariance \(\Sigma_{\mathrm{tot}}/2\).
In finite samples, \(L_n-d-1\) may replace \(L_n\) to account for the
fitted regression.

\noindent\textbf{Confidence region.}
Before inverting the Hessian estimate, project its eigenvalues onto
\([\widehat\kappa_n,L_H]\), with a positive vanishing lower floor
and a valid public upper bound as specified in
Assumption~\ref{ass:pair-design}.
Floor the eigenvalues of \(\widehat\Sigma_n^{\mathrm{raw}}\) at
\(\varepsilon_n\), or at \(\max(\varepsilon_n,\sigma^2)\) under finite
Gaussian privacy, with \(\varepsilon_n>0\) tending to zero.  These operations
give positive-definite matrices \(\widehat H_n^+\) and
\(\widehat\Sigma_n^+\).  The estimated covariance of \(\bar x_K\) in its
tangent frame is \(\widehat\Xi_n/n\), where
{
\[
 \widehat\Xi_n=\left(\widehat H_n^+\right)^{-1}
              \widehat\Sigma_n^+\left(\widehat H_n^+\right)^{-1}.
\]
}
An optional finite-sample adjustment increases the covariance when
the Hessian regression is imprecise.  Using only fitted quantities, set
\begin{align}
 u_n&=\frac{\left\{\tfrac12\tr\left(\widehat\Sigma_n^+\right)
       \tr\left[(Q_n+\lambda_nI_d)^{-1}\right]\right\}^{1/2}}
       {\max\left(\left\|\widehat H_n^+\right\|_{\mathrm F},\widehat\kappa_n\right)},\notag\\
 \widetilde\Xi_n&=\{1+a_{\rm g}\min(u_n,c)\}\widehat\Xi_n,
       \qquad a_{\rm g},c\ge0.                              \label{eq:wald-guard}
\end{align}
The constants \(a_{\rm g},c\) are fixed in advance.  The multiplier
tends to one as regression information grows, and \(a_{\rm g}=0\) gives
the unadjusted sandwich estimate.  To form a confidence region, express a
candidate point \(x\) relative to \(\bar x_K\) as
\(V_n(x)=B_{\bar x_K}^{-1}R_{\bar x_K}^{-1}(x)\).  At significance level
\(\alpha_{\rm sig}\), retain the points in \(K_{\rm s}\) satisfying
{
\begin{equation}\label{eq:wald-region}
 \mathcal C_{1-\alpha_{\rm sig},n}=\left\{x\in K_{\rm s}:
 nV_n(x)^{\mathsf T}\widetilde\Xi_n^{-1}V_n(x)
 \le\chi^2_{d,1-\alpha_{\rm sig}}\right\},
 \qquad \alpha_{\rm sig}\in(0,1).
\end{equation}
}
This is the usual covariance ellipsoid in the final tangent frame,
mapped back to the manifold and restricted to \(K_{\rm s}\).
The tangent ellipsoid has squared semi-axis lengths equal to the
eigenvalues of \(\widetilde\Xi_n/n\) multiplied by the chi-squared
quantile.  Theorem~\ref{thm:practical-wald} establishes its coverage.

\begin{algorithm}[!htbp]

\caption{Private Riemannian estimation with SPR}\label{alg:spr}
\begin{spacing}{1.05}
\begin{algorithmic}[1]
\Require Initial point \(x_0=\bar x_0\), retraction, transport, public frame,
gains \(\gamma_k\), radii \(h_k\), direction cycle, local randomiser,
step cap \(\delta_0\), metric projection \(\Pi\), regression window,
ridge and spectral bounds, and significance level \(\alpha_{\rm sig}\).
\For{each pair \(k=1,\ldots,K\)}
\State Freeze \(x_{k-1}\); construct \(q_{k,+},q_{k,-}\) by
\eqref{eq:pair-queries} and assign them to two fresh participants.
\State Receive one private gradient per participant; transport and form
\(U_k,D_k,X_k\) by \eqref{eq:pair-messages}.
\State Cap \(U_k\), update \(x_k\), and average using \eqref{eq:pair-update}.
\State After burn-in, retain \(D_k,X_k\) for SPR over complete cycles.
\EndFor
\State Fit \(\widehat H_n^+\) and \(\widehat\Sigma_n^+\) using
\eqref{eq:pair-regression}--\eqref{eq:pair-covariance}; form
\(\widetilde\Xi_n\) by \eqref{eq:wald-guard}.
\State \Return \(\bar x_K\), covariance \(\widetilde\Xi_n/n\), and
confidence region \eqref{eq:wald-region}.
\end{algorithmic}
\end{spacing}
\end{algorithm}

\subsection{Assumptions}\label{sec:assumptions}

The following conditions concern identification on the public basin,
the released scores, population smoothness, and tuning.

\begin{assumption}[Local geometry and identification]\label{ass:local}
The public compact sets satisfy
\(K_{\rm s}\subset\operatorname{int}(K_{\rm q}),\quad K_{\rm q}\subset U\), where
\(K_{\rm s}\) is geodesically convex and \(U\) is a normal convex
neighbourhood.  The geometric maps used by the algorithm, including the
retraction inverses and parallel transports, are \(C^3\) on neighbourhoods
of their compact active domains.  There are
\(x_\star\in\operatorname{int}(K_{\rm s})\) and \(\mu>0\) such that
\(\nabla F(x_\star)=0\) and

\begin{equation}\label{eq:local-monotonicity}
 \left\langle-\Log_x(x_\star),\nabla F(x)\right\rangle_x
 \ge\mu d(x,x_\star)^2,\qquad x\in K_{\rm s}.
\end{equation}

\end{assumption}

\begin{assumption}[Released scores]\label{ass:released-score}
Users' data and fresh randomisation seeds are independent across users;
the data distributions may vary with the user index.  For
\(i\in\{2k-1,2k\}\) and every admissible public query \(q\),

\[
 \E\{\widehat g_i(q)\mid\mathcal G_{k-1}\}=\nabla F(q),\qquad
 \sup_{i,q}\E\{\norm{\widehat g_i(q)}^4\mid\mathcal G_{k-1}\}
 \le C\quad\text{a.s.}
\]

The conditional covariance fields \(C_i(x,q)\) are continuous in the
base and query points, uniformly in the user index.  Along every admissible
predictable sequence with user index tending to infinity and
\((x,q)\to_P(x_\star,x_\star)\),

\[
 C_i(x,q)\to_P\Sigma_{\mathrm{tot}},
\]

where \(\Sigma_{\mathrm{tot}}\) is deterministic and positive semidefinite.
The continuity and convergence are uniform over fixed-horizon rows.
\end{assumption}

\begin{assumption}[Population smoothness]\label{ass:pair-smooth}
The objective \(F\) is twice continuously differentiable on an open
neighbourhood of \(K_{\rm q}\), and its Riemannian Hessian is locally
Lipschitz under parallel transport.
\end{assumption}

\begin{assumption}[Tuning]\label{ass:pair-design}\label{ass:iterates}
Let \(\gamma_k=\gamma_0(k+k_0)^{-\alpha}\), where
\(\gamma_0>0\), \(k_0\ge0\), and \(1/2<\alpha<1\).
In the fixed-horizon regime, \(h_k=h_n\) and

\begin{equation}\label{eq:radius-window}
 h_n\to0,\qquad \sqrt n\,h_n\to\infty,\qquad
 \sqrt n\,h_n^2\to0.
\end{equation}

In the online regime, \(h_k\asymp k^{-\beta}\) after finitely many
cycles, with \(1/4<\beta<1/2\).  The nuisance window discards \(o(K)\)
initial pairs, and \(\lambda_n/s_n\to0\).
The transcript-measurable Hessian floor satisfies
\(0<\widehat\kappa_n<L_H\) and \(\widehat\kappa_n\to_P0\), where the
public upper bound \(L_H\) exceeds \(\lambda_{\max}(H)\) for
\(H=\nabla^2F(x_\star)\).  All bounds, neighbourhoods, and covariance
conditions above hold uniformly throughout each fixed-horizon row,
including initialization.
\end{assumption}

\begin{remark}
The public basin encodes prior localisation information; its projection
does not use \(x_\star\).  The single-point monotonicity condition and
population smoothness imply \(H\succeq\mu I_d\) and the local mean-field
expansions used below.  Conditional fourth moments and Lipschitz Hessians
also appear in analyses of averaged stochastic approximation
\citep{Tripuraneni2018}; here the state moment rates are derived from the
public safeguard.  For fixed-horizon calculations, write
\(K_n=n/2\), \(h_{n,k}=h_n\), and, when needed,
\(x_{n,k},\mathcal G_{n,k}\); these row indices are otherwise suppressed.
The power rule \(h_n=cn^{-\beta}\), \(1/4<\beta<1/2\), satisfies
\eqref{eq:radius-window}, balancing query bias and Hessian information.

An independently verified fixed floor
\(0<\kappa<\lambda_{\min}(H)\) may replace the vanishing Hessian floor.
If no valid fixed upper bound is available, the upper endpoint may instead
be a deterministic sequence \(L_{H,n}\to\infty\), with
\(0<\widehat\kappa_n<L_{H,n}\); the same consistency argument applies.
For the additive Gaussian mechanism, total covariance gives
\(\Sigma_{\mathrm{tot}}=\Sigma_0+\sigma^2I_d\), where \(\Sigma_0\) is the
limiting covariance of the unrandomised score.  Positive definiteness of
\(\Sigma_{\mathrm{tot}}\) is needed only for nondegenerate Wald inference.
\end{remark}

The projection and cap are part of the reported algorithm.  They are public
post-processing of the released transcript and do not alter user-level LDP.
The next theorem establishes the state rates used below.

\begin{theorem}
\label{thm:primitive-pair-stability}
Suppose Assumptions~\ref{ass:local}--\ref{ass:pair-design} hold, and put $
 \Delta_k=R_{x_\star}^{-1}(x_k).$ Then,
\begin{equation}\label{eq:pair-stability}
 \E\norm{\Delta_k}^2=O(\gamma_k+h_k^4),\qquad
 \E\norm{\Delta_k}^4=O(\gamma_k^2+h_k^8).
\end{equation}
Consequently \(x_k\to_Px_\star\) and
\(q_{k,+},q_{k,-}\to_Px_\star\).  If
\(m_K=\lfloor K^\kappa\rfloor\) for any \(0<\kappa<1/2\), then, for every
fixed \(\rho>0\) such that \(B_{3\rho}(x_\star)\subset\operatorname{int}(K_{\rm s})\),
\begin{equation}\label{eq:tail-localisation}
 \Pr\!\left\{\max_{m_K\le j\le K}d(x_j,x_\star)\ge\rho\right\}\to0,
\end{equation}
and
\begin{equation}\label{eq:tail-safeguard-inactive}
 \Pr\{\Pi(y_j)=y_j,\ \widetilde U_j=U_j
       \text{ for all }m_K<j\le K\}\to1.
\end{equation}
 Thus the first \(m_K=o(\sqrt K)\) projected updates are asymptotically
negligible in the reported average, while the inferential tail follows the
ordinary local recursion with probability tending to one.
\end{theorem}

The bound \(O(\gamma_k)\) is the usual stochastic-approximation fluctuation,
while \(h_k^4\) is the squared bias introduced by the symmetric query pair.
The first term parallels stability results for stochastic algorithms on
manifolds \citep{Pelletier1998,Bonnabel2013,Tripuraneni2018}; the second records
the additional price of probing away from the current iterate.  The tail
localisation statement connects those local analyses to the implemented
algorithm: the public cap and projection guarantee a well-defined recursion,
but disappear from the root-\(n\) expansion once the iterates enter the local
basin.

\section{Theoretical results}\label{sec:theory}

The stability bounds control the state error and the symmetric-query bias.
We now derive the averaged recursion's limiting distribution and show how
the same message pairs estimate its covariance for feasible inference.

\subsection{Asymptotic normality}\label{sec:private-clt}

The following expansion displays the two roles of a message pair.  Write
\[
 m_{k,\pm}=\E(z_{k,\pm}\mid\mathcal G_{k-1}),\qquad
 \varepsilon_{k,\pm}=z_{k,\pm}-m_{k,\pm}.
\]

\begin{lemma}\label{lem:pair-expansions}
Under Assumptions~\ref{ass:local}--\ref{ass:pair-smooth},
\begin{align}
 \E(U_k\mid\mathcal G_{k-1})
 &=B_{x_{k-1}}^{-1}\nabla F(x_{k-1})+b_k,
&\norm{b_k}&\le Ch_k^2,\label{eq:proof-U-mean}\\
 \E(D_k\mid\mathcal G_{k-1})
 &=H_{k-1}X_k+r_k,
&\norm{r_k}&\le h_k\omega(h_k),\label{eq:proof-D-mean}
\end{align}
where \(H_{k-1}=D\Psi_{x_{k-1}}(0)\) and one may take \(\omega(h)=Ch\).  Moreover,
\[
 \xi_k=U_k-\E(U_k\mid\mathcal G_{k-1}),\qquad
 \zeta_k=D_k-\E(D_k\mid\mathcal G_{k-1})
\]
are martingale differences.  If
\(\Omega_{k,\pm}=\Cov(z_{k,\pm}\mid\mathcal G_{k-1})\), then
\begin{equation}\label{eq:proof-pair-covariances}
 \Cov(\xi_k\mid\mathcal G_{k-1})
 =\Cov(\zeta_k\mid\mathcal G_{k-1})
 =\frac14(\Omega_{k,+}+\Omega_{k,-}).
\end{equation}
\end{lemma}

The average equation has a second-order query bias, while the difference
equation is a first-order regression for the Hessian.  The exact covariance
identity supplies the factor two in \eqref{eq:pair-covariance} and makes the
point and nuisance calculations compatible on the user scale.

\begin{theorem}[Averaged-RSGD limit]
\label{thm:private-clt}
Under Assumptions~\ref{ass:local}--\ref{ass:pair-design}, with \(n=2K\),

\begin{equation}\label{eq:private-clt}
 \sqrt n\,R_{x_\star}^{-1}(\bar x_K)
 \xRightarrow[]{D}\Normal(0,\Xi),\qquad
 \Xi=H^{-1}\Sigma_{\mathrm{tot}}H^{-1}.
\end{equation}

\end{theorem}

\begin{corollary}\label{cor:gauss-inflation}
For the tangent Gaussian mechanism \eqref{eq:tangent-gauss},
\[
 \Xi=H^{-1}(\Sigma_0+\sigma^2I_d)H^{-1}.
\]
\end{corollary}

Without privacy noise, Theorem~\ref{thm:private-clt} reduces to the familiar
Hessian-sandwich limit for averaged Riemannian stochastic approximation
\citep{Tripuraneni2018}.  The same sandwich form governs batch intrinsic means
\citep{BhattacharyaPatrangenaru2003,BhattacharyaPatrangenaru2005}; here it is
obtained from an online recursion whose observations are never revealed.  A
conditionally centred release adds its covariance to \(\Sigma_0\) but does not
move the first-order equation.  Moreover, pairing two users halves the
variance of a pair-average update, so conversion from \(K\) pairs to
\(n=2K\) users restores the first-order covariance \(\Xi/n\).  This is the
same user-scale normalisation as an ordinary one-message averaged recursion,
rather than the effective sample size of a holdout estimator.

\subsection{Hessian and covariance from the same messages}
\label{sec:theory-inference}

The same pair transcript identifies both nuisance matrices.  At completed
cycles,

\begin{equation}\label{eq:pair-gram}
 Q_n=\frac{s_n}{d}I_d.
\end{equation}

Curvature identification therefore does not depend on random excursions of
the optimization path.

\begin{theorem}[Transcript nuisance consistency]
\label{thm:H-consistency}\label{thm:sigma-consistency}
Under Assumptions~\ref{ass:local}--\ref{ass:pair-design},
\[
 \widehat H_n^{\mathrm{raw}}\xrightarrow[]{P}H,\qquad
 \widehat\Sigma_n^{\mathrm{raw}}\xrightarrow[]{P}\Sigma_{\mathrm{tot}}.
\]
The spectral projection of the Hessian and any covariance projection whose
lower floor vanishes, or equals the known Gaussian-noise floor \(\sigma^2\),
preserve these limits.
\end{theorem}

The two nuisance estimators use the messages that also drove the point
recursion.  Predictability and martingale laws of large numbers replace any
independence between point and nuisance estimation.

Theorem~\ref{thm:H-consistency} supplies the step that an asymptotic law with
population \(H\) and \(\Sigma_{\mathrm{tot}}\) does not provide: it makes the
sandwich covariance computable from the released stream.  Controlled
perturbations have long been used to recover derivatives in stochastic
approximation \citep{Ruppert1985,Spall2000}.  In the present construction, the
completed direction cycles make \(Q_n\) exactly isotropic, and the symmetric
message differences turn those perturbations into a full Hessian regression
in a common tangent frame.  Thus Hessian identification is separated from
the accidental directions explored by the optimization path.  This point is
also relevant to Euclidean online private inference, where uncertainty has
otherwise been obtained through problem-specific plug-in or self-normalising
constructions \citep{Liu2023,XieEtAl2025,CaiHuSunWu2025,CaiHuWu2026Federated}.

\begin{theorem}[Intrinsic Wald inference]
\label{thm:practical-wald}
Suppose Assumptions~\ref{ass:local}--\ref{ass:pair-design} hold and
\(\Sigma_{\mathrm{tot}}\succ0\).  Then

\begin{equation}\label{eq:Xi-practical-consistency}
 \widehat\Xi_n\xrightarrow[]{P}\Xi,
 \qquad \widetilde\Xi_n\xrightarrow[]{P}\Xi
\end{equation}

in the smooth public frame at the final average.  For the tangent
coordinates \(V_n(x)\) defined in Section~\ref{sec:private-procedure},

\begin{equation}\label{eq:wald-stat}
 T_n=nV_n(x_\star)^{\mathsf T}\widetilde\Xi_n^{-1}V_n(x_\star)
 \xRightarrow[]{D}\chi_d^2,
\end{equation}

and the region \eqref{eq:wald-region} has asymptotic coverage
\(1-\alpha_{\rm sig}\).
\end{theorem}

The Wald conclusion follows from nuisance consistency and the usual
studentisation argument for smooth \(M\)-estimators
\citep{wpVanderVaart1998}.  Its substantive content is that every term in the
studentiser is measurable from public queries and one locally private message
per user.  The covariance estimate accounts jointly for sampling variation
and randomisation variation, while the Hessian estimate converts that score
variation into uncertainty on the manifold.  The resulting region therefore
has the same local quadratic form as classical intrinsic-mean inference
\citep{BhattacharyaPatrangenaru2005}, but requires neither the raw observations
nor a population Hessian.

\subsection{Special cases and extensions}\label{sec:theory-extensions}

\begin{corollary}\label{cor:smooth-functional}
Let \(h:M\to\R^r\) be differentiable at \(x_\star\), with $
 J_h=D(h\circ\Exp_{x_\star})(0),
 \Sigma_h=J_h\Xi J_h^{\mathsf T}.
$
Under the conditions of Theorem~\ref{thm:practical-wald},
\[
 \sqrt n\{h(\bar x_K)-h(x_\star)\}
 \xRightarrow[]{D}\Normal(0,\Sigma_h).
\]
If \(\Sigma_h\) is positive definite and a public
\(\widehat J_{h,n}\) satisfies \(\widehat J_{h,n}\to_P J_h\), then
\(\widehat\Sigma_{h,n}=\widehat J_{h,n}\widetilde\Xi_n
\widehat J_{h,n}^{\mathsf T}\) is consistent and
\[
 n\{h(\bar x_K)-h(x_\star)\}^{\mathsf T}
 \widehat\Sigma_{h,n}^{-1}
 \{h(\bar x_K)-h(x_\star)\}
 \xRightarrow[]{D}\chi_r^2.
\]
\end{corollary}

This corollary turns the intrinsic covariance estimate into inference for
smooth scalar and vector summaries, including coordinates, contrasts, and
smooth functions of eigenvectors or positive-definite matrices.  The
calculation is post-processing of the released transcript.

\begin{proposition}\label{prop:euclidean-specialization}
Let \(M=\R^d\) with the standard metric, use ordinary addition, identity
transport, and a constant frame.  Algorithm~1 then becomes an ordinary
Polyak--Ruppert recursion whose pair differences are regressed on
\(h_kv_k\) by Euclidean central differences.  Under the Euclidean versions of
Assumptions~\ref{ass:local}--\ref{ass:pair-design},
\[
 \sqrt n(\bar\theta_K-\theta_\star)
 \Rightarrow \Normal(0,H^{-1}\Sigma_{\mathrm{tot}}H^{-1}),
\]
and Theorems~\ref{thm:H-consistency} and \ref{thm:practical-wald} hold with
ordinary vector differences.  Curvature, moving-frame, retraction, and
transport remainders vanish identically.
\end{proposition}

\begin{corollary}\label{cor:fixed-radius-affine}
Retain Assumptions~\ref{ass:local}--\ref{ass:pair-design}, except for the shrinking-radius conditions.
If the transported mean score is affine with constant derivative \(H\), the
symmetric finite difference has no higher-order bias and a fixed feasible
radius \(h>0\) may be used.  Suppose that, for every direction phase
\(r=1,\ldots,R\) and sign \(s\in\{+,-\}\), the conditional covariance of an
individual transported release at the corresponding query converges to
\(\Sigma_{r,s}\), and define
\[
 \Sigma_h=\frac1{2R}\sum_{r=1}^R
                 (\Sigma_{r,+}+\Sigma_{r,-}).
\]
Then Theorems~\ref{thm:private-clt}--\ref{thm:practical-wald} hold with
\(\Sigma_{\mathrm{tot}}\) replaced by \(\Sigma_h\), provided
\(\Sigma_h\succ0\) for the Wald conclusion.  The original covariance is
recovered when \(\Sigma_h=\Sigma_{\mathrm{tot}}\); a covariance field that
is constant on a neighbourhood containing the target and the finite query
orbit is sufficient.
\end{corollary}

The scale in \eqref{eq:wald-stat} is the original number of users.  The pair
construction therefore imposes deterministic Hessian excitation without a
holdout sample or a second message from any participant.

Proposition~\ref{prop:euclidean-specialization} also locates the source of the
difficulty.  Estimating the Hessian from a single stochastic-gradient path can
be poorly conditioned even in \(\R^d\); manifold geometry adds transport,
retraction, and moving-frame remainders but does not create the excitation
problem.  Corollary~\ref{cor:fixed-radius-affine} further shows that a shrinking
query radius is needed to control nonlinear finite-difference bias, not as a
consequence of local privacy itself.

\begin{proposition}
\label{prop:public-preconditioner}
Let \(P_x:T_xM\to T_xM\) be a public, predictable, smooth field of
self-adjoint maps whose eigenvalues lie in a fixed interval
\([p_-,p_+]\subset(0,\infty)\).  Modify only the point update in
Algorithm~1 by replacing \(U_k\) with \(P_{x_{k-1}}U_k\) before applying the
step cap.  Suppose Assumptions~\ref{ass:local}--\ref{ass:pair-design} hold,
except that \eqref{eq:local-monotonicity} is replaced by
\[
 \langle-\Log_x(x_\star),P_x\nabla F(x)\rangle_x
 \ge \mu_P d(x,x_\star)^2.
\]
Let \(\phi(x)=R_{x_\star}^{-1}(x)\) in the limiting frame and
\(J(x)=D_0[u\mapsto\phi\{R_x(B_xu)\}]\).
Assume also, writing \(P_\star\) in that frame,
\[
 J(x)B_x^{-1}P_x\nabla F(x)
 =P_\star H R_{x_\star}^{-1}(x)+O\{d(x,x_\star)^2\}.
\]
Then Theorems~\ref{thm:private-clt}--\ref{thm:practical-wald} continue to
hold with the same sandwich covariance $
 \Xi=H^{-1}\Sigma_{\mathrm{tot}}H^{-1}.
$
The Hessian and covariance estimators continue to use the unpreconditioned
pair messages.
\end{proposition}

At the target, preconditioning changes the linearised drift from \(H\) to
\(P_\star H\) and the pair-average noise covariance from
\(\Sigma_{\mathrm{tot}}/2\) to
\(P_\star\Sigma_{\mathrm{tot}}P_\star/2\).  These factors cancel in the
Polyak--Ruppert sandwich covariance.  The pair regression remains on the
original score scale, so the same transcript nuisance estimators apply.  The
proof is given in the Supplementary Material.

Appendix~\ref{app:strengthened-results} gives two further robustness results.  The first controls
target bias and covariance distortion when unbounded scores with finite
moments are clipped at a growing radius.  The second shows that, at a fixed
clipping radius, the private sandwich covariance converges to its nonprivate
counterpart as the randomisation noise vanishes.

\section{Statistical applications}\label{sec:model-classes}

The framework covers several statistical targets for which the parameter is
naturally manifold valued.  Directional location and von Mises--Fisher models
are standard tools for unit-vector data \citep{MardiaJupp2000,Banerjee2005};
principal components are directions rather than unrestricted Euclidean
parameters \citep{JolliffeCadima2016}; Fr\'echet means extend averages to
metric and Riemannian spaces \citep{Frechet1948,Pennec2006}; and
positive-definite matrices arise as covariance and diffusion-tensor summaries
\citep{Bhatia2009,Arsigny2006,Arsigny2007}.  The following six examples match
the numerical designs A--F.  Appendix~S.1 gives the
full model formulas and sufficient local conditions.  Their derivations
are given in the Supplementary Material.

\subsection{Directional location on the sphere}

Let \(Y\in\sph^{p-1}\) and estimate a direction
\(\mu\in\sph^{p-1}\) with the von Mises--Fisher loss
\(\ell(\mu,Y)=-\kappa\mu^{\mathsf T}Y\).  Its tangent score is
\[
 g(\mu,Y)=-\kappa(I-\mu\mu^{\mathsf T})Y,
 \qquad \norm{g(\mu,Y)}\le\kappa.
\]
If \(m=\E(Y)\ne0\), the target is \(\mu_\star=m/\norm m\) and the population
Hessian is the scalar operator
\(H=\kappa\norm m I_{T_{\mu_\star}\sph^{p-1}}\).  Hence concentration of the
directional distribution supplies local identification, while the bounded
score gives a direct privacy calibration.  Design A uses this model on
\(\sph^2\).

\begin{corollary}\label{cor:vmf}
Under Assumptions~\ref{ass:local}--\ref{ass:pair-design} and \ref{ass:ex-vmf}, the conclusions of
Theorems~\ref{thm:private-clt}--
\ref{thm:practical-wald} hold.  With tangent Gaussian privacy,
\[
 \sqrt n\,R_{\mu_\star}^{-1}(\bar x_K)
 \Rightarrow \Normal(0,\Xi_{\rm dir}),\qquad
 \Xi_{\rm dir}=\frac{\kappa^2\Cov(P_\star Y)+\sigma^2I_{p-1}}
 {\kappa^2\norm m^2},
 \quad P_\star=I-\mu_\star\mu_\star^{\mathsf T}.
\]
For an exact von Mises--Fisher model, this becomes an isotropic covariance on
the tangent space.  The same private transcript consistently estimates
\(\Xi_{\rm dir}\) and yields the Wald region in
\eqref{eq:wald-region}.
\end{corollary}

\subsection{Streaming principal component analysis}

For a bounded vector \(Y\in\R^p\), the leading eigenvector of
\(\E(YY^{\mathsf T})\) minimises
\(\ell(u,Y)=-(u^{\mathsf T}Y)^2/2\) over \(u\in\sph^{p-1}\).  The score
\[
 g(u,Y)=-(I-uu^{\mathsf T})YY^{\mathsf T}u
\]
has norm at most \(B^2/2\) when \(\norm Y\le B\).  In the tangent eigenbasis,
the Hessian eigenvalues are \(\lambda_1-\lambda_j\), \(j=2,\ldots,p\), so the
usual eigengap is exactly the local identification condition.  This example
is a standard test case for averaged Riemannian stochastic approximation
\citep{Tripuraneni2018}; Design B evaluates its locally private inferential
version on \(\sph^3\).

\begin{corollary}\label{cor:pca}
Let \(Y_j=u_j^{\mathsf T}Y\) in the population eigenbasis and set
$
 H_{\rm pca}=\diag(\lambda_1-\lambda_2,\ldots,\lambda_1-\lambda_p),
 \Sigma_{\rm pca}=\Cov\{(Y_1Y_2,\ldots,Y_1Y_p)^{\mathsf T}\}.
$
Under Assumptions~\ref{ass:local}--\ref{ass:pair-design} and \ref{ass:ex-pca},
Theorems~\ref{thm:private-clt}--\ref{thm:practical-wald} hold with
\[
 \Xi_{\rm pca}=H_{\rm pca}^{-1}
   (\Sigma_{\rm pca}+\sigma^2I_{p-1})H_{\rm pca}^{-1},
\]
where the tangent Gaussian mechanism is calibrated with substitution
sensitivity \(B^2\).  Thus the usual eigengap controls both local
identification and the amplification of private score noise.
\end{corollary}

\subsection{Log-Euclidean means of positive-definite matrices}

Under the log-Euclidean metric, \(X\mapsto\log X\) identifies \(\SPD_m\) with
the Euclidean space of symmetric matrices.  Writing \(S=\log X\) and
\(T=\log Y\), the loss and score are
\[
 \ell(S,T)=\tfrac12\norm{S-T}_{\mathrm F}^2,
 \qquad g(S,T)=S-T.
\]
The target is \(S_\star=\E(T)\), the Hessian is the identity, and the score
covariance is \(\Cov(T)\).  Bounded or publicly clipped log coordinates give
finite sensitivity.  Because the score is affine, Design C can use the
fixed-radius regime in Corollary~\ref{cor:fixed-radius-affine}, although its
implementation still estimates the identity Hessian from private releases.

\begin{corollary}\label{cor:logeuc}
Under Assumptions~\ref{ass:local}--\ref{ass:pair-design} and \ref{ass:ex-log},
\[
 \sqrt n\{\log(\bar x_K)-\E(\log Y)\}
 \Rightarrow
 \Normal\left\{0,\Cov(\log Y)+\sigma^2I\right\}.
\]
Here the general sandwich covariance reduces to the released-score
covariance because \(H=I\).  Corollary~\ref{cor:fixed-radius-affine} permits
a fixed query radius, and the transcript estimator still gives a feasible
Wald region without supplying this known Hessian to the algorithm.
\end{corollary}

\subsection{Intrinsic Fr\'echet means}

For observations on a Hadamard manifold, the intrinsic mean minimises
\(F(x)=\E\{d(x,Y)^2\}/2\), with score
\[
 g(x,Y)=-\Log_x(Y).
\]
If the public query region and data support lie in bounded sets, then
\(\norm{g(x,Y)}\le D\), where \(D\) is their maximum geodesic separation.
The population Hessian is the expected Hessian of squared distance.  Normal
convex neighbourhoods and a positive Fr\'echet Hessian are the standard local
conditions for uniqueness and asymptotic inference
\citep{BhattacharyaPatrangenaru2005,Afsari2011}.  Design D uses the
hyperbolic plane, where the absence of a cut locus makes the local geometric
conditions particularly transparent.

\begin{corollary}\label{cor:frechet}
Under Assumptions~\ref{ass:local}--\ref{ass:pair-design} and \ref{ass:ex-frechet},
\[
 \sqrt n\,R_{x_\star}^{-1}(\bar x_K)
 \Rightarrow
 \Normal\!\left[0,
 H^{-1}\left\{\E\bigl(\Log_{x_\star}(Y)\ot\Log_{x_\star}(Y)\bigr)
             +\sigma^2I\right\}H^{-1}\right].
\]
Thus the private limit retains the classical intrinsic-mean sandwich form,
with the tangent randomisation variance added to the log-map covariance.
The transcript estimates both matrices in this expression.
\end{corollary}

\subsection{Robust transformed Fr\'echet centres}

A smooth radial transform replaces squared distance by
\(\ell(x,Y)=\psi\{d(x,Y)^2/2\}\) and gives the weighted score
\[
 g(x,Y)=-\psi'\{d(x,Y)^2/2\}\Log_x(Y).
\]
For the pseudo-Huber transform, the weight decreases with distance and the
score norm is bounded by its scale parameter.  The population Hessian combines
the Hessian of squared distance with a rank-one term involving \(\psi''\).
Assumption~\ref{ass:ex-transformed} keeps this
operator positive definite even though \(\psi''<0\).  Design E therefore
examines robust location under both geometric curvature and heterogeneous radial
weights.

\begin{corollary}\label{cor:tfrechet}
Let \(H_\psi\) be the population Hessian described above.  Under
Assumptions~\ref{ass:local}--\ref{ass:pair-design} and \ref{ass:ex-transformed},
\[
 \begin{aligned}
 \sqrt n\,R_{x_\star}^{-1}(\bar x_K)
   &\Rightarrow \Normal(0,\Xi_\psi),
 \end{aligned}
\]
where $ \Xi_\psi
   =H_\psi^{-1}\Bigl(
      \E\!\left[\psi'\{d(x_\star,Y)^2/2\}^2
      \Log_{x_\star}(Y)\ot\Log_{x_\star}(Y)\right]
      +\sigma^2I\Bigr)H_\psi^{-1}.$ The same conclusion therefore covers the pseudo-Huber centre used in
Design E, with privacy calibrated by its bounded weighted score.
\end{corollary}

\subsection{Affine-invariant covariance estimation}

On \(\SPD_m\) with the affine-invariant metric, the Gaussian covariance
quasi-likelihood
\[
 \ell(X,Y)=\tfrac12\log\det X+\tfrac12Y^{\mathsf T}X^{-1}Y
\]
has Riemannian score \(g(X,Y)=(X-YY^{\mathsf T})/2\).  Its target is
\(X_\star=\E(YY^{\mathsf T})\), and in affine-invariant normal coordinates the
Hessian equals \(I/2\).  Restricting \(X\) to a compact spectral band and
bounding or clipping \(Y\) controls sensitivity.  Design F studies the
resulting three-dimensional local score for \(2\times2\) covariance matrices.

\begin{corollary}\label{cor:spd-like}
Write \(\xi=X_\star^{-1/2}Y\) and use affine-invariant normal coordinates at
\(X_\star\).  Under Assumptions~\ref{ass:local}--\ref{ass:pair-design} and
\ref{ass:ex-spd},
\[
 \sqrt n\,R_{X_\star}^{-1}(\bar x_K)
 \Rightarrow \Normal(0,\Xi_{\rm spd}),\qquad
 \Xi_{\rm spd}=\Cov(\xi\xi^{\mathsf T})+4\sigma^2I_{\Sym(m)}.
\]
The factor four is the direct specialization of the sandwich formula to
\(H=I/2\).  The private transcript estimates this covariance and supports the
intrinsic Wald region without using the known-form Hessian.
\end{corollary}

Across the six examples, the inferential ingredients have the same form: a
positive local Hessian identifies the target, a bounded or clipped tangent
score calibrates the local randomiser, and SPR
estimates both the Hessian and the released-score covariance.

\section{Numerical experiments}\label{sec:sim}
The simulation study evaluates the complete no-holdout procedure in six
models.  Design A is directional location on \(\sph^2\), B is bounded
streaming PCA on \(\sph^3\), C is a log-Euclidean mean on \(\SPD_2\), D is an
intrinsic Fr\'echet mean on \(\HH^2\), E is a pseudo-Huber center on
\(\HH^2\), and F is an affine score in three local coordinates of an
  \(\SPD_2\) covariance model.  Appendix~S.3 gives the
  data-generating models and error metrics.

We use \(n\in\{4000,8000,16000,32000\}\),
\(\eps\in\{\infty,8,4,2,1\}\), and \(\del=10^{-6}\).  Every
design--privacy--sample-size cell contains 500 independent repetitions.
A calibration stage with separate random seeds selected the gain, radius,
and finite-sample guard, after which the settings were frozen.  The nonlinear
models use a decreasing radius with \(1/4<\beta<1/2\); the affine score fields
in C and F use Corollary~\ref{cor:fixed-radius-affine}.

At each checkpoint, the private pair transcript supplies the Hessian,
released-score covariance, and Wald region, with the statistic scaled by the
original number of users \(n\).  Appendix~S.3 records
the complete data-generating models, tuning, safeguards, initialisation,
error metrics, and comparator specifications.

\begin{figure}[!htbp]
\centering
\includegraphics[width=.96\textwidth,height=.72\textheight,keepaspectratio]
{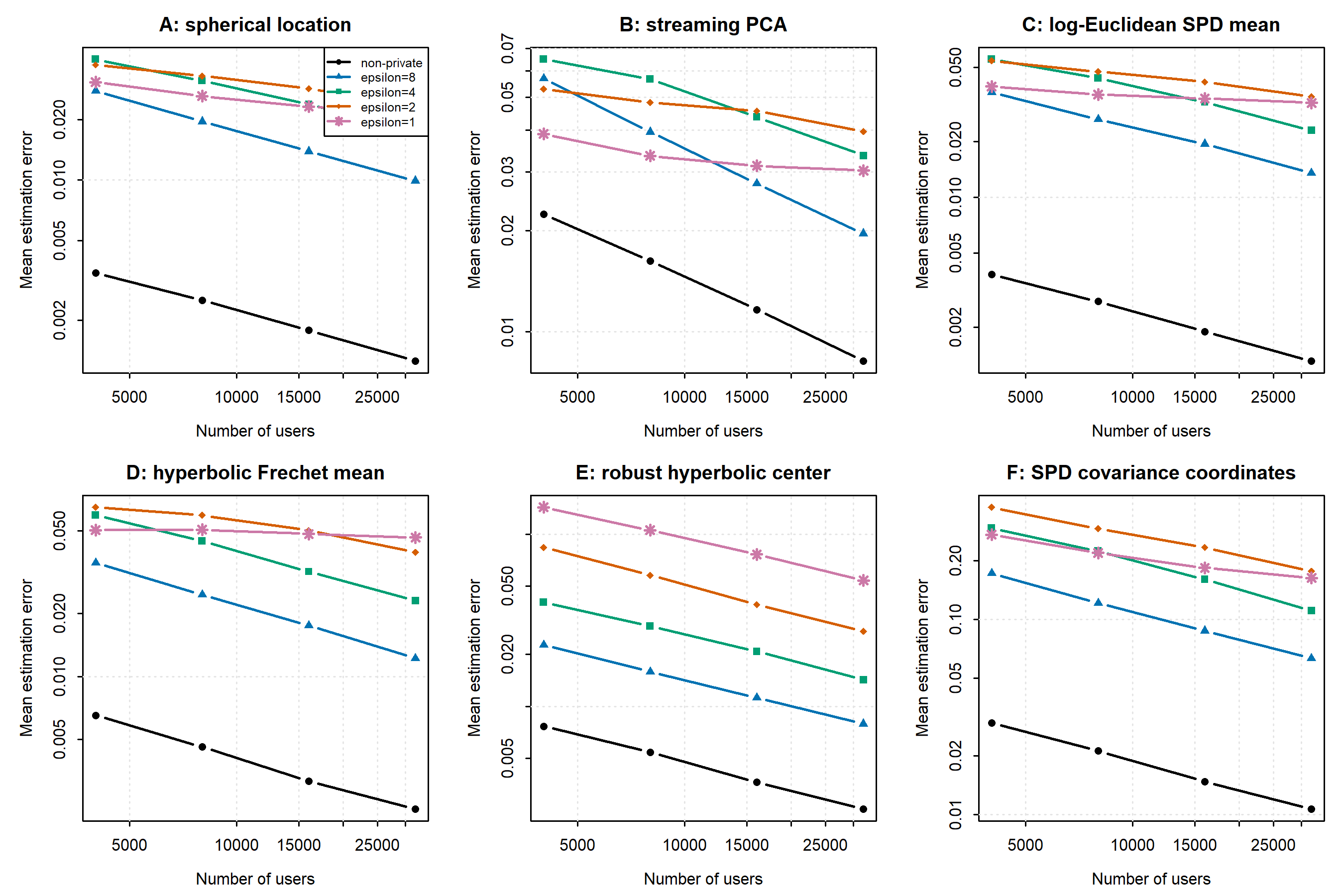}
\caption{Mean estimation error of the SPR procedure in
designs A--F.  Error is geodesic distance in A, D, and E, sign-aligned angle
in B, and local coordinate distance in C and F.  Each point is based on 500
repetitions.}
\label{fig:sim-error-af}
\end{figure}

Figure~\ref{fig:sim-error-af} shows that terminal estimation error at
\(n=32000\) is below its value at \(n=4000\) in every design--privacy
curve.  The privacy cost is most visible in the three-dimensional models B,
C, and F.

\begin{figure}[!htbp]
\centering
\includegraphics[width=.96\textwidth,height=.72\textheight,keepaspectratio]
{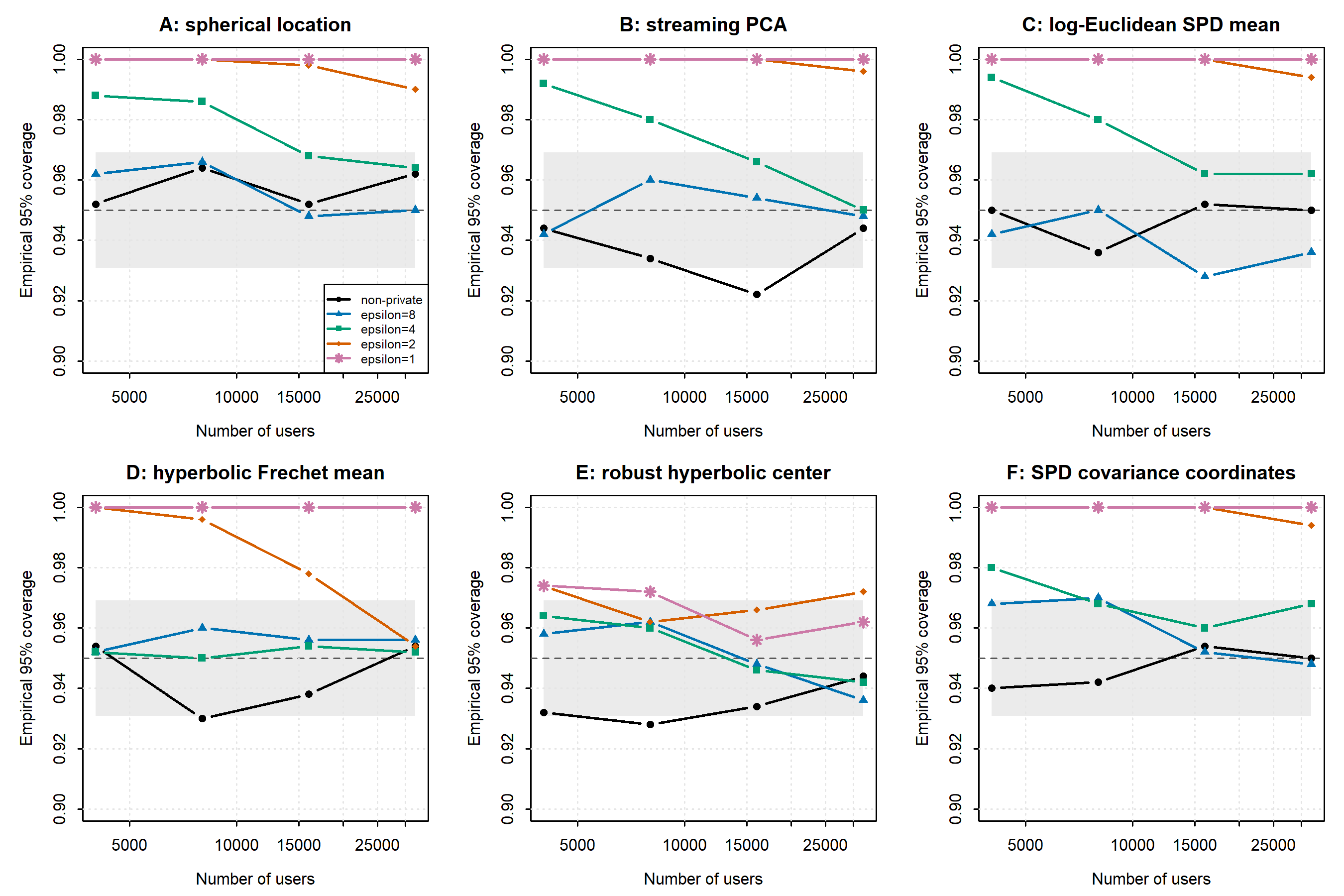}
\caption{Empirical coverage of nominal 95\% Wald regions for the no-holdout
SPR procedure in designs A--F.  Each point is based on 500
repetitions; the grey band is
\(0.95\pm2\sqrt{0.95(0.05)/500}\).}
\label{fig:sim-coverage-af}
\end{figure}

Figure~\ref{fig:sim-coverage-af} shows that, for \(\eps=8\), terminal coverage
ranges from 0.936 to 0.956, and for
\(\eps=4\) it ranges from 0.942 to 0.968.  The nonprivate terminal range is
0.944--0.962.  At \(\eps\in\{2,1\}\), several regions become conservative
as privacy noise widens the estimated uncertainty.
Appendix~S.3 reports the terminal error and
coverage for every design and privacy level.

\subsection{Comparison with two alternative approaches}
\label{sec:sim-comparators}

Both alternatives begin with the ordinary one-query recursion: participant
\(i\) receives the current public iterate, returns one private tangent score,
and the server updates and averages the iterate.  The gain exponent is 0.60
and \(C=(0.18,0.24,0.60,0.75,0.75,1.00)\) in designs A--F.  To bound the
first private update, trajectory regression uses
$ \gamma_i=C(i+b_\eps)^{-0.60},
 b_\eps=\max\{0,(C\sigma/c_\eps)^{1/0.60}-1\}, $
where \(c_\eps\) is a public noise-step cap.  The half-sample recursion uses
\(\gamma_i=C_\eps i^{-0.60}\), with
\(C_\eps=\min(C,c_\eps/\sigma)\) under finite privacy and \(C_\eps=C\)
without privacy; design F instead uses the offset form above.  At
\(\eps=(\infty,8,4,2,1)\), the caps are respectively
\((.85,.85,.85,.85,.85)\) in A and D,
\((1.35,1.60,1.35,1.35,1.35)\) in B,
\((.75,.75,.75,.75,.75)\) in C,
\((1,1,2,1.5,2)\) in E, and
\((.85,.85,.10,.20,.40)\) in F.  The privacy mechanism, public working
region, score bound, and data-generating model are otherwise the same as for
the SPR procedure.

\paragraph{Same-trajectory regression.}
At a checkpoint \(n\), the recursive average after \(\lfloor n/3\rfloor\)
updates is frozen as a public anchor \(a_n\).  For
\(\mathcal J_n=\{\lfloor n/3\rfloor+1,\ldots,n\}\), the later query point and
its released score are transported to \(T_{a_n}M\) and expressed in the
public frame as \(w_i\) and \(z_i\).  With \(\bar w_n\) and \(\bar z_n\)
denoting their averages, the fitted slope is
\[
 \begin{aligned}
 Q_n^{\rm tr}
   =\sum_{i\in\mathcal J_n}(w_i-\bar w_n)(w_i-\bar w_n)^{\mathsf T},\;
 A_n^{\rm tr}
   =\sum_{i\in\mathcal J_n}(z_i-\bar z_n)(w_i-\bar w_n)^{\mathsf T},\;
 \widehat B_n^{\rm tr}
   =A_n^{\rm tr}(Q_n^{\rm tr}+\lambda_nI)^{-1}.
 \end{aligned}
\]
The Hessian estimate is the symmetric part of \(\widehat B_n^{\rm tr}\),
after projecting its eigenvalues onto \([n^{-1/4},L_H]\).  If
\(r_i=z_i-\bar z_n-\widehat B_n^{\rm tr}(w_i-\bar w_n)\), the released-score
covariance is the positive-semidefinite projection of
\[
 \frac{1}{|\mathcal J_n|-d-1}\sum_{i\in\mathcal J_n}r_ir_i^{\mathsf T}.
\]
We use
\(\lambda_n=n^{-1/2}\{1+\tr(Q_n^{\rm tr})/d\}\), covariance floor
\(\min(10^{-8},n^{-1})\), and public upper bounds
\(L_H=(5,5.6,1,1.4/\tanh(1.4),1,0.6)\) for A--F.  The resulting sandwich
covariance is paired with the all-user average and the Wald statistic is
scaled by \(n\).

\paragraph{Half-sample public probes.}
Let \(m=\lfloor n/2\rfloor\).  The first \(m\) participants form the averaged
private point estimate \(a_n\); the remaining \(n-m\) participants are queried
at public offsets that cycle through
\(e_1,\ldots,e_d,-e_1,\ldots,-e_d\) around \(a_n\).  Each participant again
returns one private score.  After transport to \(T_{a_n}M\), the actual query
displacements and released scores are inserted into the same centred
regression and residual-covariance formulas above; denote the resulting
centred Gram matrix by \(Q_n^{\rm pr}\).  Here
\(\lambda_n=m^{-1/2}\{1+\tr(Q_n^{\rm pr})/d\}\), the Hessian floor is
\(n^{-1/4}\), the covariance floor is \(\min(10^{-8},m^{-1})\), and the same
upper bounds \(L_H\) are used.  The probe radius is \(0.65n^{-0.025}\) in
A--C and F and \(0.85n^{-0.025}\) in D and E; for the nonprivate E cell its
exponent is 0.15.  Because only the first block estimates the target, its
Wald statistic is scaled by \(m\), not by \(n\).

These implementations use only public queries and released data.  They can
also yield asymptotically valid Wald regions: trajectory regression requires
persistent excitation of the realised path, whereas half-sample probes
require a shrinking radius and increasing probe information.  Appendix~S.3.2 states both results; their proofs are given in the
Supplementary Material.  The comparison places
side by side three complete ways of acquiring Hessian information: one
relies on random path variation, the second pays the efficiency cost of sample
splitting, and the SPR method builds balanced excitation into the
all-user recursion.  All three implementations use the frozen tuning and
finite-sample safeguards recorded in Appendix~S.3.

\begin{figure}[!htbp]
\centering
\includegraphics[width=.87\textwidth,height=.50\textheight,keepaspectratio]
{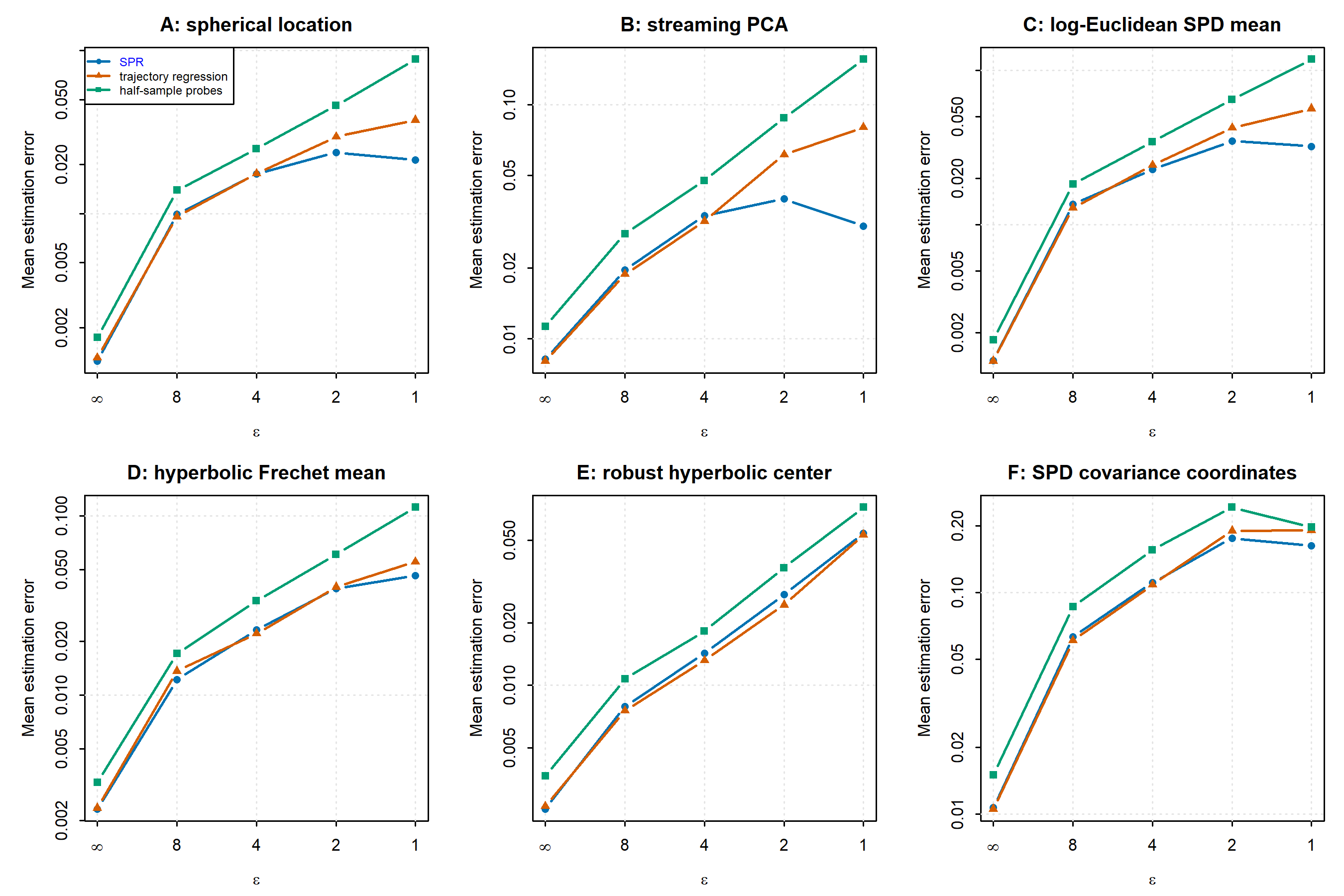}
\caption{Terminal mean estimation error at \(n=32000\) for the proposed
SPR method, same-trajectory regression, and half-sample public
probes.  Each point is based on 500 repetitions.}
\label{fig:sim-method-error-af}
\end{figure}

At \(n=32000\), the half-sample construction has larger point error than the
SPR procedure in every displayed setting
(Figure~\ref{fig:sim-method-error-af}).  Same-trajectory regression has
similar error in several weak-privacy cells.  At \(\eps=1\), the
SPR error is smaller than both alternatives in A--D and F and is
within 1.5\% of trajectory regression in E.

\begin{figure}[!htbp]
\centering
\includegraphics[width=.87\textwidth,height=.50\textheight,keepaspectratio]
{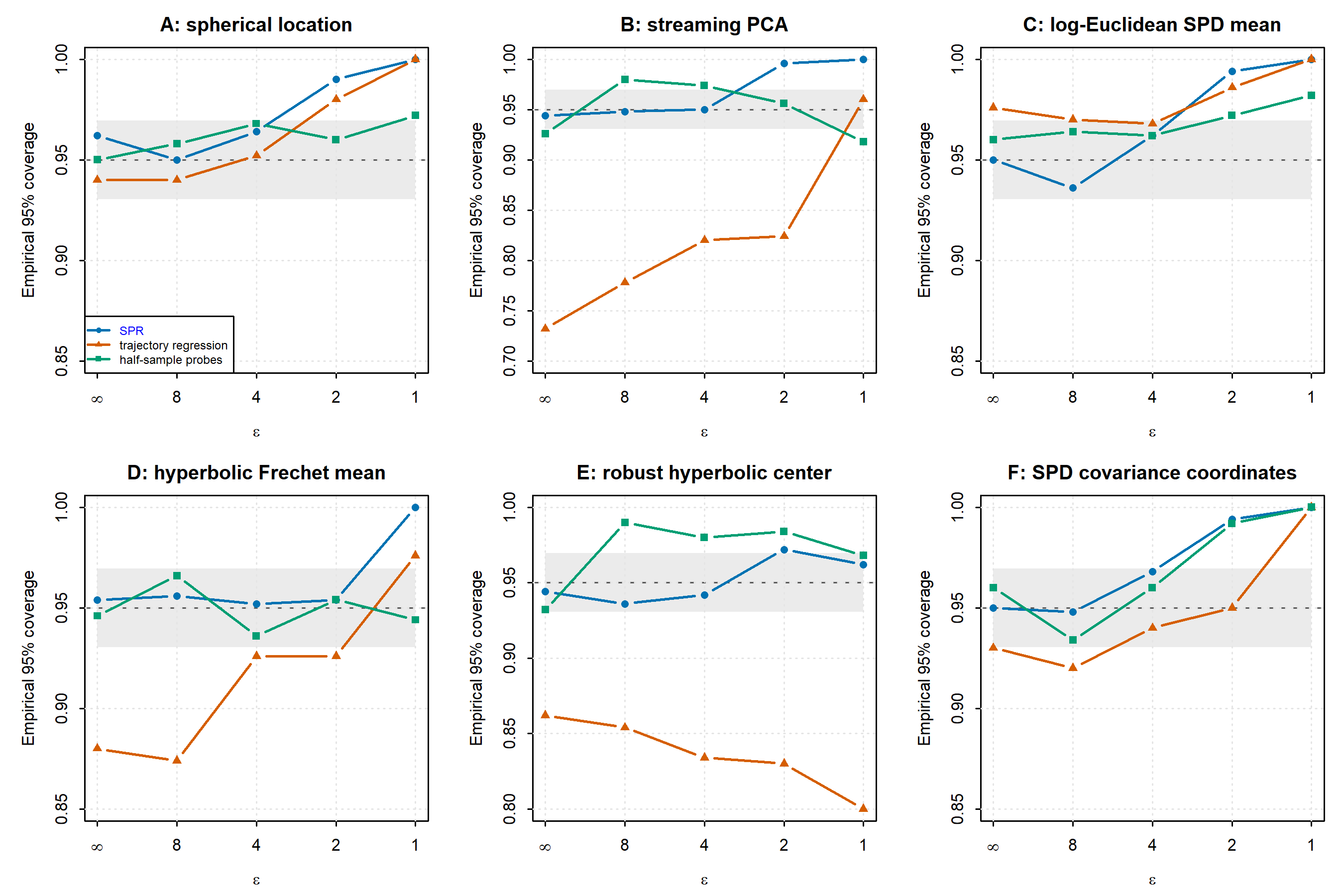}
\caption{Terminal empirical coverage of nominal 95\% Wald regions at
\(n=32000\) for the proposed SPR method, same-trajectory
regression, and half-sample public probes.  Each point is based on 500
repetitions; the grey band is
\(0.95\pm2\sqrt{0.95(0.05)/500}\).}
\label{fig:sim-method-coverage-af}
\end{figure}

The coverage comparison reveals the difference between point accuracy and
usable studentisation (Figure~\ref{fig:sim-method-coverage-af}).  Trajectory
regression falls to 0.732 in B and 0.800 in E, consistent with inadequate
finite-sample excitation of some Hessian directions.  Balanced public probes
restore coverage in most settings, but incur the point-estimation cost shown
in Figure~\ref{fig:sim-method-error-af}.  SPR combines controlled
excitation with an all-user point recursion: coverage ranges from 0.936 to 1
over the 30 settings and is at least 0.962 in all six models at \(\eps=1\).

\section{Anthropometric data}\label{sec:nhanes}

The National Health and Nutrition Examination Survey (NHANES) records
standardised body measurements obtained by trained technicians.  These
measurements are used to monitor overweight and obesity and to study the
relation between body size and health \citep{CDCBodyMeasures2020}.  We analyse
standing height, body weight, and waist circumference among examined adults,
using the public microdata to emulate the local-release design.
Their leading joint direction summarizes overall body size and central
adiposity; waist relative to height is also widely used in cardiometabolic
screening \citep{Ashwell2012}.

The 2015--2016 cycle provides public preprocessing information.
Among 5,359 adults with complete measurements, it fixes the centre and
scale of each variable, the orientation, and a positive-definite update
preconditioner.  In the public tangent frame, the preconditioner is the
inverse of the earlier-cycle eigengap Hessian.  It is applied to the pair
average inside the step cap; the unpreconditioned pair differences
continue to estimate the Hessian and released-score covariance.
For Hessian truncation, use the vanishing floor
\(\widehat\kappa_n=\min\{G,\tfrac12\widehat s_n n^{-1/4}\}\), where
\(\widehat s_n=\max\{1,\tfrac12\sum_{j=1}^2
|\lambda_j(\widehat H_n^{\rm raw})|\}\) and \(G=B^2/2\),
with upper endpoint \(2G\).  The floor is active for the
\(\eps=2,1\) fits; each reported sandwich uses the corresponding
truncated Hessian.  Proposition~\ref{prop:public-preconditioner} gives the
conditions under which fixed public preconditioning preserves the
first-order sandwich covariance.  The analysis sample contains 5,175 complete adult records
from 2017--2018.  After standardisation by the earlier-cycle moments, each
vector is clipped to Euclidean norm \(B=2.5\).  The clipping fractions are
10.3\% in the preprocessing cycle and 11.6\% in the analysis cycle.  The
analysis targets the leading direction of the clipped second moment under
the unweighted distribution of examined adults satisfying these criteria;
the NHANES sampling design is described in \citet{CDCDesign2020}.  We report
nominal model-based Wald regions for the independent-user design emulation.
The batch
2017--2018 eigenvector is used to report angular error.

At public query \(u\), a clipped vector \(y\) contributes the tangent score
\[
 g(u,y)=-(I-uu^{\mathsf T})yy^{\mathsf T}u.
\]
Its norm is bounded by \(B^2/2\), giving substitution sensitivity \(B^2\).
Each participant releases one tangent Gaussian message with
\(\del=10^{-6}\).  The first 5,174 participants form 2,587 symmetric pairs;
the final participant contributes one ordinary update.  Pair averages drive
the point recursion, while 2,568 post-burn-in differences over completed
cycles estimate the Hessian and released-score covariance.  The observed
excitation Gram matrix has condition number one at every privacy level.
Uncertainty is computed with the full analysis count \(n=5175\).

\begin{table}[ht]
\centering
\caption{NHANES anthropometric principal component analysis
under local privacy.  Angular error is measured against the batch non-private target.
The last column is the largest semi-axis of the nominal model-based 95\% Wald
region for the independent-user design emulation.}
\label{tab:nhanes-pca}

\begin{tabular}{rrrrrr}
\toprule
\(\epsilon\) & Height & Weight & Waist & Angle error (\(^{\circ}\)) &
95\% semi-axis (\(^{\circ}\)) \\
\midrule
\(\infty\) & 0.409 & 0.670 & 0.619 & 0.26 & 2.61 \\
8 & 0.450 & 0.656 & 0.606 & 2.34 & 19.75 \\
4 & 0.407 & 0.641 & 0.651 & 2.60 & 50.23 \\
2 & 0.380 & 0.637 & 0.671 & 4.11 & 82.98 \\
1 & 0.379 & 0.648 & 0.661 & 3.42 & 86.30 \\
\bottomrule
\end{tabular}

\end{table}

\newcommand{\NHANESApplicationCaption}{%
Anthropometric principal-direction estimates and uncertainty in the NHANES
local-release emulation. (a) Reported loadings for height, weight and waist
circumference; dotted lines mark the corresponding batch loadings.
(b) Angular difference from the 2017--2018 batch reference.
(c) Largest semi-axis of the nominal model-based 95\% Wald region.
Panels (b) and (c) use different vertical scales; semi-axes are not error bars
for the angular differences. The five fits use 5,175 complete adult records,
public preprocessing from 2015--2016, norm clipping at $B=2.5$ and
$\delta=10^{-6}$. The target is the leading direction of the unweighted,
standardised and clipped second moment. Privacy levels are categorical;
$\epsilon=\infty$ denotes the nonprivate streaming fit, not the batch reference.
The Hessian uses a vanishing eigenvalue floor.
The regions concern the independent-user protocol emulation, rather than
design-based inference for the complex survey.}

\begin{figure}[!htbp]
  \centering
  \includegraphics[width=\textwidth]{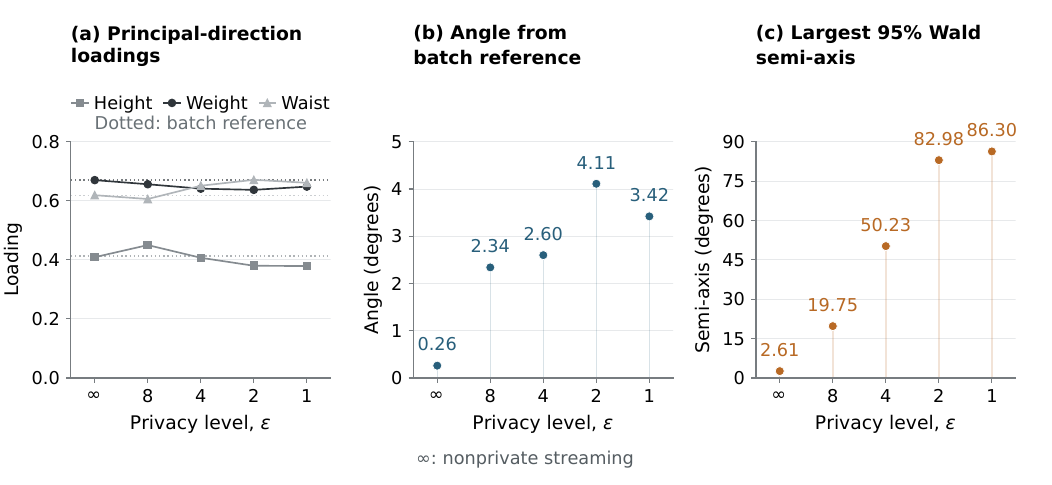}
  \caption{\NHANESApplicationCaption}
  \label{fig:nhanes-application-results}
\end{figure}

Table~\ref{tab:nhanes-pca} reports the private estimates and uncertainty.
Figure~\ref{fig:nhanes-application-results} compares the loadings, angular differences and uncertainty across privacy levels.  The
batch reference has standardised loadings
\((0.413,0.670,0.617)\) for height, weight, and waist circumference.  The
non-private streaming estimate is within \(0.26^\circ\) of this direction.
At \(\eps=8\) and \(\eps=4\), the private estimates are within
\(2.34^\circ\) and \(2.60^\circ\); their loadings retain the dominant
weight--waist pattern.  Their nominal 95\% major semi-axes, measured as angular
radii after the normalizing retraction, are \(19.75^\circ\) and
\(50.23^\circ\).  At \(\eps=2\) and \(\eps=1\), the point directions
remain similar, while the semi-axes widen to \(82.98^\circ\) and
\(86.30^\circ\).  The point estimates retain the leading anthropometric
pattern in these runs, but the broad regions give little directional
precision under stronger privacy.

\section{Concluding remarks}\label{sec:conclusion}
Locally private tangent gradients preserve the first-order equation of a
manifold-valued target, and SPR turns the
same one-pass stream into both an all-user Polyak-Ruppert estimator and a
Hessian regression.  Pair averages drive optimization, pair differences
identify the full Hessian and released-score covariance, and the resulting
Wald region uses only public queries and released messages.  The theory
retains the root-\(n\) scale based on the original number of users.  Across
six simulation models, error decreases with sample size and coverage is
near nominal under moderate privacy; the NHANES analysis illustrates
private estimation of an anthropometric principal direction and its
uncertainty.

The same construction suggests two extensions.  An adaptive public direction
cycle could concentrate probes in directions with imprecisely estimated Hessian action
while retaining a lower bound on the design Gram matrix.  Extending the
method to nonsmooth geodesically convex losses would instead require a
replacement for the central Hessian regression, such as a smoothed score or
a locally regularised generalized derivative.  Both extensions preserve the
one-message-per-user architecture that motivates the present procedure and
remain directions for future work.

\section*{Acknowledgment}
\par  Qirui Hu's research was supported by National Natural Science Foundation of China (NSFC) (Grant Nos.12601520), the Shanghai Engineering Research Center of Finance Intelligence (Grant No.~19DZ2254600) and by TRR 391 \textit{Spatio-temporal Statistics for the Transition of Energy and Transport} (Project number 520388526) funded by the Deutsche Forschungsgemeinschaft (DFG, German Research Foundation).

\clearpage
\appendix
\setcounter{section}{0}
\setcounter{equation}{0}
\setcounter{figure}{0}
\setcounter{table}{0}
\setcounter{theorem}{0}
\setcounter{proposition}{0}
\setcounter{corollary}{0}
\setcounter{lemma}{0}
\setcounter{definition}{0}
\setcounter{assumption}{0}
\renewcommand{\thesection}{S.\arabic{section}}
\renewcommand{\thesubsection}{S.\arabic{section}.\arabic{subsection}}
\renewcommand{\theequation}{S.\arabic{equation}}
\renewcommand{\thefigure}{S.\arabic{figure}}
\renewcommand{\thetable}{S.\arabic{table}}
\renewcommand{\thetheorem}{S.\arabic{theorem}}
\renewcommand{\theproposition}{S.\arabic{proposition}}
\renewcommand{\thecorollary}{S.\arabic{corollary}}
\renewcommand{\thelemma}{S.\arabic{lemma}}
\renewcommand{\thedefinition}{S.\arabic{definition}}
\renewcommand{\theassumption}{S.\arabic{assumption}}
\renewcommand{\theHsection}{S.\arabic{section}}
\renewcommand{\theHequation}{S.\arabic{equation}}
\renewcommand{\theHfigure}{S.\arabic{figure}}
\renewcommand{\theHtable}{S.\arabic{table}}
\renewcommand{\theHtheorem}{S.\arabic{theorem}}
\renewcommand{\theHproposition}{S.\arabic{proposition}}
\renewcommand{\theHcorollary}{S.\arabic{corollary}}
\renewcommand{\theHlemma}{S.\arabic{lemma}}
\renewcommand{\theHdefinition}{S.\arabic{definition}}
\renewcommand{\theHassumption}{S.\arabic{assumption}}
\renewcommand{\theremark}{S.\arabic{remark}}
\section{Model-specific formulas and conditions}\label{app:model-examples}

Each example below records the objective, stochastic gradient, sensitivity,
Hessian, and covariance used in the corresponding corollary of the main
text.  Derivations are given in the Supplementary Material.

\subsection{Intrinsic Fr\'echet means on Hadamard manifolds}\label{sec:ex-frechet}

\begingroup
\begin{assumption}\label{ass:ex-frechet}
Let $M$ be a Hadamard manifold.  Let $K_{\rm s}$ be a public compact closed
geodesically convex set containing $x_\star$ in its interior, let
$K_{\rm q}$ contain a positive-radius tube around $K_{\rm s}$, and suppose
$Y\in K_Y$ almost surely for a compact set $K_Y$.
\end{assumption}
\endgroup
Consider
\begin{equation*}
        \ell(x,y)=\frac12d(x,y)^2,
        \qquad
        F(x)=\frac12\E[d(x,Y)^2].
\end{equation*}
The stochastic gradient is
\begin{equation*}
        g(x,Y)=-\Log_x(Y),
        \qquad
        \norm{g(x,Y)}_x=d(x,Y)\le D,
        \qquad D:=\sup_{x\in K_{\rm q},y\in K_Y}d(x,y).
\end{equation*}
Thus $G=D$ for LDP calibration. Let $\mu=x_\star$ be the intrinsic mean. The first-order condition is
\begin{equation*}
        \E[\Log_\mu(Y)]=0.
\end{equation*}
Therefore
\begin{equation}\label{eq:frechet-sigma}
        \Sigma_0=
        \E[\Log_\mu(Y)\ot\Log_\mu(Y)] .
\end{equation}
The Hessian is
\begin{equation}\label{eq:frechet-H}
        H=
        \E\!\left[\left.\Hess_x\left(\frac12d(x,Y)^2\right)\right|_{x=\mu}\right].
\end{equation}
On a Hadamard manifold there is no cut locus, and
$x\mapsto d(x,y)^2/2$ is geodesically 1-strongly convex.  Compactness of
$K_{\rm q}\times K_Y$ gives the required derivative bounds.  Metric
projection onto $K_{\rm s}$ is target-quasi-nonexpansive, so
Assumption~\ref{ass:local} holds with $\mu=1$ when the initialization lies in
$K_{\rm s}$.

\subsection{Smooth transformed Fr\'echet means and pseudo-Huber centers}\label{sec:ex-transformed}

\begingroup
\begin{assumption}\label{ass:ex-transformed}
Use the manifold, public sets, and data support in
Assumption~\ref{ass:ex-frechet}.  The transform $\psi$ is $C^3$ on
$[0,D^2/2]$, where $D=\sup_{x\in K_{\rm q},y\in K_Y}d(x,y)$, and
\[
c_\psi:=\inf_{0\le t\le D^2/2}
\{\psi'(t)+2t\min(\psi''(t),0)\}>0.
\]
\end{assumption}
\endgroup
Consider the loss
\begin{equation*}
        \ell(x,y)=\psi\!\left(\frac12d(x,y)^2\right).
\end{equation*}
The gradient is
\begin{equation}\label{eq:transformed-grad}
        g(x,Y)=
        -\psi'\!\left(\frac12d(x,Y)^2\right)\Log_x(Y),
        \qquad
        \norm{g(x,Y)}_x\le G_\psi:=D\sup_{0\le t\le D^2/2}|\psi'(t)|.
\end{equation}
The Hessian of the sample loss is
\begin{equation}\label{eq:transformed-hess}
\begin{split}
        \Hess_x\ell(x,y)
        ={}&\psi'\!\left(\frac12d(x,y)^2\right)
        \Hess_x\left(\frac12d(x,y)^2\right) \\
        &+\psi''\!\left(\frac12d(x,y)^2\right)
        \Log_x(y)\ot\Log_x(y).
\end{split}
\end{equation}
Assumption~\ref{ass:ex-transformed} ensures local strong convexity while
allowing $\psi''<0$.  It is satisfied by the pseudo-Huber transform
\begin{equation*}
        \psi_\rho(t)=\rho^2\left(\sqrt{1+2t/\rho^2}-1\right),
\end{equation*}
because $\psi_\rho'(t)+2t\psi_\rho''(t)=(1+2t/\rho^2)^{-3/2}>0$.

At the minimiser $x_\star$,
\begin{equation*}
        \E\left[\psi'\!\left(\frac12d(x_\star,Y)^2\right)\Log_{x_\star}(Y)\right]=0.
\end{equation*}
Thus
\begin{equation}\label{eq:tfrechet-sigma}
        \Sigma_0=
        \E\!\left[
        \psi'\!\left(\frac12d(x_\star,Y)^2\right)^2
        \Log_{x_\star}(Y)\ot\Log_{x_\star}(Y)
        \right],
\end{equation}
and
\begin{equation}\label{eq:tfrechet-H}
\begin{split}
        H=\E\!\Bigg[&
        \psi'\!\left(\frac12d(x_\star,Y)^2\right)
        \left.\Hess_x\left(\frac12d(x,Y)^2\right)\right|_{x=x_\star} \\
        &+\psi''\!\left(\frac12d(x_\star,Y)^2\right)
        \Log_{x_\star}(Y)\ot\Log_{x_\star}(Y)
        \Bigg].
\end{split}
\end{equation}

\subsection{Directional location and von Mises--Fisher likelihood on the sphere}\label{sec:ex-vmf}

\begingroup
\begin{assumption}\label{ass:ex-vmf}
Let $M=\sph^{p-1}$, $Y\in\sph^{p-1}$, $\kappa>0$, and
$m=\E Y\ne0$.  The public stable and query sets are contained in a
geodesic ball of radius $R<\pi/2$ about $m/\norm m$.
\end{assumption}
\endgroup
For known $\kappa$, the negative von Mises--Fisher log-likelihood up to constants is
\begin{equation*}
        \ell(\mu,y)=-\kappa\mu^\top y,
        \qquad \mu\in\sph^{p-1}.
\end{equation*}
The Riemannian gradient is the tangent projection of the Euclidean gradient:
\begin{equation*}
        g(\mu,Y)=-\kappa(I-\mu\mu^\top)Y,
        \qquad
        \norm{g(\mu,Y)}\le\kappa.
\end{equation*}
Thus $G=\kappa$. The minimiser is
\begin{equation*}
        \mu_\star=\frac{m}{\norm m}.
\end{equation*}
Writing $P_\star=I-\mu_\star\mu_\star^\top$, the covariance of the stochastic gradient is
\begin{equation}\label{eq:vmf-sigma}
        \Sigma_0=\kappa^2P_\star\E[YY^\top]P_\star\big|_{T_{\mu_\star}\sph^{p-1}}
        =\kappa^2\Cov(P_\star Y).
\end{equation}
The Hessian is the scalar tangent operator
\begin{equation}\label{eq:vmf-hess}
        H=\kappa\norm m\,I_{T_{\mu_\star}\sph^{p-1}}.
\end{equation}
The stability condition is also explicit.  If
\(\theta=d(\mu,\mu_\star)\le R<\pi/2\), then
\[
 \left\langle-\Log_\mu(\mu_\star),\nabla F(\mu)\right\rangle
 =\kappa\norm m\,\theta\sin\theta
 \ge \kappa\norm m\frac{\sin R}{R}\,\theta^2.
\]
Thus Assumption~\ref{ass:local} holds whenever the public stable and query
sets lie in such a cap.  For nested radii \(0<R_{\rm s}<R_{\rm q}<\pi/2\),
the exponential retraction and deterministic query choice
\(h_k\le R_{\rm q}-R_{\rm s}\) keep both symmetric queries in the
larger cap; the public metric projection, or an alternative safeguard satisfying Section~\ref{app:alternative-safeguards}, keeps
the iterates in the smaller one.
If $Y$ is exactly von Mises--Fisher with concentration $\kappa$ and direction $\mu_\star$, then $\norm m=A_p(\kappa)$ and rotational symmetry gives
\begin{equation*}
        \Sigma_0=\kappa A_p(\kappa)I_{T_{\mu_\star}\sph^{p-1}},
\end{equation*}
where $A_p(\kappa)$ is the usual mean resultant length.

\subsection{Leading eigenvector estimation on the sphere}\label{sec:ex-pca}

\begingroup
\begin{assumption}\label{ass:ex-pca}
Let $Y\in\R^p$, $\norm Y\le B$ almost surely, and let
$\Sigma=\E[YY^\top]$ have eigenvalues $\lambda_1>\lambda_2\ge\cdots\ge\lambda_p$.
A public reference $c\in\sph^{p-1}$ fixes the sign of the leading
eigenvector $u_1$ through $c^\top u_1\ge\eta>0$.
The public stable and query sets lie in a cap of radius $R<\pi/2$
about $u_1$ and select this representative.
\end{assumption}
\endgroup
Consider
\begin{equation*}
        \ell(u,y)=-\frac12(u^\top y)^2,
        \qquad u\in\sph^{p-1}.
\end{equation*}
The gradient is
\begin{equation*}
        g(u,Y)=-(I-uu^\top)YY^\top u,
        \qquad
        \norm{g(u,Y)}\le \frac12B^2.
\end{equation*}
Thus one may take $G=B^2/2$, with substitution sensitivity $B^2$.
In the tangent eigenbasis
$\{u_2,\ldots,u_p\}$,
\begin{equation}\label{eq:pca-hess}
        H=\diag(\lambda_1-\lambda_2,\ldots,\lambda_1-\lambda_p).
\end{equation}
Writing $Y_j:=u_j^\top Y$, the gradient at the optimum has tangent coordinates
\begin{equation*}
        g(u_1,Y)=-(Y_1Y_2,\ldots,Y_1Y_p)^\top,
\end{equation*}
so
\begin{equation}\label{eq:pca-sigma}
        \Sigma_0=\Cov\{(Y_1Y_2,\ldots,Y_1Y_p)^\top\}.
\end{equation}

The eigengap gives a direct basin condition.  Write
\(u=\cos\theta\,u_1+\sin\theta\,w\), where
\(w\perp u_1\) and \(\|w\|=1\).  For
\(0\le\theta\le R<\pi/2\),
\[
 \left\langle-\Log_u(u_1),\nabla F(u)\right\rangle
 =\{\lambda_1-w^\top\Sigma w\}\theta\sin\theta\cos\theta
 \ge (\lambda_1-\lambda_2)
       \frac{\sin R\cos R}{R}\,\theta^2.
\]
Consequently, Assumption~\ref{ass:local} holds on any public stable and
query sets contained in this cap.  Nested cap radii
\(0<R_{\rm s}<R_{\rm q}<\pi/2\), together with
the exponential retraction and \(h_k\le R_{\rm q}-R_{\rm s}\), verify
symmetric-query feasibility.
The public sign reference ensures that the selected cap contains only the
desired representative of the leading eigendirection.

For a Gaussian PCA benchmark,
$\Sigma_0=\lambda_1\diag(\lambda_2,\ldots,\lambda_p)$ in the eigenbasis.  The
locally private implementation uses norm-bounded observations.  With bounded
support this is immediate; when raw observations are clipped, the population
target is the leading eigenvector of the corresponding clipped second moment.

\subsection{Affine-invariant SPD covariance quasi-likelihood}\label{sec:ex-spd-likelihood}

Let $M=\SPD_m$ be the cone of $m\times m$ symmetric positive-definite matrices with affine-invariant metric
\begin{equation*}
        \ip{U}{V}_X=\tr(X^{-1}UX^{-1}V).
\end{equation*}
For $y\in\R^m$, consider the Gaussian covariance negative log-likelihood
\begin{equation*}
        \ell(X,y)=\frac12\log\det X+\frac12y^\top X^{-1}y.
\end{equation*}
The Riemannian gradient is
\begin{equation}\label{eq:spd-like-grad}
        g(X,Y)=\frac12(X-YY^\top).
\end{equation}
\begingroup
\begin{assumption}\label{ass:ex-spd}
The query matrices lie in the compact spectral band
\begin{equation*}
        K=\{X\in\SPD_m:aI\preceq X\preceq bI\},
\end{equation*}
with $0<a<b<\infty$, $\norm Y\le B$ almost surely, and
$X_\star=\E[YY^\top]\in\operatorname{int}(K)$.
\end{assumption}
\endgroup
Then
\begin{equation*}
        \norm{g(X,Y)}_X
        =\frac12\norm{I-X^{-1/2}YY^\top X^{-1/2}}_F
        \le\frac12\left(\sqrt m+\frac{B^2}{a}\right),
\end{equation*}
so
\begin{equation*}
        G=\frac12\left(\sqrt m+\frac{B^2}{a}\right).
\end{equation*}
The population target is $X_\star=\E[YY^\top]$, provided $X_\star\in K$. In affine-invariant normal coordinates at $X_\star$, $U\leftrightarrow W=X_\star^{-1/2}UX_\star^{-1/2}$, the Hessian is
\begin{equation*}
        H=\frac12I_{\Sym(m)}.
\end{equation*}
Let $\xi=X_\star^{-1/2}Y$. The gradient at $X_\star$ has coordinates
\begin{equation*}
        \widetilde g(X_\star,Y)=\frac12(I-\xi\xi^\top),
\end{equation*}
and hence
\begin{equation}\label{eq:spd-like-sigma}
        \Sigma_0=\frac14\Cov(\xi\xi^\top)
\end{equation}
as an operator on $\Sym(m)$ with the Frobenius inner product.

The quasi-likelihood derivatives remain valid for Gaussian covariance models. The locally private implementation uses bounded or clipped observations to obtain finite global sensitivity.

\subsection{Log-Euclidean SPD means}\label{sec:ex-log-euclidean}

Let $M=\SPD_m$ with the log-Euclidean metric. The map $X\mapsto S=\log X$ is an isometry from $(\SPD_m,\text{log-Euclidean})$ to the Euclidean vector space $\Sym(m)$ with the Frobenius inner product. For a log-Euclidean mean, set
\begin{equation*}
        \ell(S,T)=\frac12\norm{S-T}_F^2,
        \qquad S=\log X,
        \qquad T=\log Y.
\end{equation*}
The target is $S_\star=\E[T]$ and $X_\star=\exp(S_\star)$. In log coordinates,
\begin{equation*}
        g(S,T)=S-T,
        \qquad H=I_{\Sym(m)},
        \qquad \Sigma_0=\Cov(T).
\end{equation*}
\begingroup
\begin{assumption}\label{ass:ex-log}
The variable $T=\log Y$ is supported in a compact set $\mathcal K_T$.
The public query log coordinates lie in a compact set $\mathcal K_S$
containing $\E T$ in its interior.
\end{assumption}
\endgroup
Then
\begin{equation*}
        G=\sup_{S\in\mathcal K_S,\,T\in\mathcal K_T}\norm{S-T}_F.
\end{equation*}

\section{Changing clipping and privacy regimes}\label{app:strengthened-results}
The preceding results use a fixed inferential regime. This section gives transfer tools for clipping levels and privacy budgets that change with sample size. The bias bound identifies when a clipped target can be replaced by the original target, and the triangular-array result standardizes any verified asymptotic linear representation by the score covariance of the current regime.

\begin{proposition}\label{prop:wp-clipping-moment}
Let \(\Psi=B_{x_\star}^{-1}g(x_\star,Y)\), assume
\(\E\Psi=0\), and suppose \(\E\|\Psi\|^q\le M_q<\infty\) for some
\(q>2\).  For the radial cap \(\mathsf C_\tau\) defined in
Section~\ref{sec:private-procedure}, let
\(m_\tau=\E\mathsf C_\tau(\Psi)\).  Then
\[
 \|m_\tau\|\le M_q\tau^{1-q},\qquad
 \left\|\E\{\mathsf C_\tau(\Psi)\mathsf C_\tau(\Psi)^{\mathsf T}\}
       -\E(\Psi\Psi^{\mathsf T})\right\|_{\mathrm{op}}
 \le 2M_q\tau^{2-q}.
\]
For \(x\) in a fixed normal neighbourhood of \(x_\star\), define the
transported mean fields
\[
 G(x)=\E\{B_{x_\star}^{-1}\Gamma_x^{x_\star}g(x,Y)\},\qquad
 G_\tau(x)=\E\mathsf C_\tau\{
 B_{x_\star}^{-1}\Gamma_x^{x_\star}g(x,Y)\}.
\]
Suppose \(G\) is continuously differentiable with nonsingular
\(DG(x_\star)=H\), \(G_\tau\) is continuously differentiable, and $
 \sup_x\|D G_\tau(x)-D G(x)\|_{\mathrm{op}}\longrightarrow0.
$
For all sufficiently large \(\tau\), the local root \(x_\tau\) of
\(G_\tau(x)=0\) exists uniquely and satisfies
\[
 \|\Log_{x_\star}(x_\tau)\|
 \le 4\|H^{-1}\|_{\mathrm{op}}M_q\tau^{1-q}.
\]
Consequently, a limit centred at \(x_{\tau_n}\) is also valid for
\(x_\star\) under any normalisation \(a_n\) satisfying
\(a_n\tau_n^{1-q}\to0\).
\end{proposition}

The proposition connects the bounded-score privacy mechanism to models with
unbounded scores and finite moments.  Increasing the clipping radius controls
both the estimating-equation bias and the covariance distortion; the final
condition makes the clipped and original targets indistinguishable on the
inferential scale.  Proofs are given in the Supplementary Material.

\begin{proposition}\label{prop:wp-nonprivate-limit}
Fix \(\tau\) and let \(x_\tau\) be the local root of \(G_\tau(x)=0\)
from Proposition~\ref{prop:wp-clipping-moment}.  Write
\(H_\tau=DG_\tau(x_\tau)\) and $
 S_\tau=\mathsf C_\tau\{
 B_{x_\star}^{-1}\Gamma_{x_\tau}^{x_\star}g(x_\tau,Y)\}.
$
Suppose the released score in the same frame can be written as
\(\xi_\eps=S_\tau+\eta_\eps\), where
\(\E(\eta_\eps\mid Y)=0\) and
\(\E\|\eta_\eps\|^2\to0\) as \(\eps\to\infty\).  Then $
 \operatorname{Var}(\xi_\eps)\longrightarrow
 \Sigma_\tau:=\operatorname{Var}(S_\tau),
$
and the private sandwich covariance converges to
$
 H_\tau^{-1}\Sigma_\tau H_\tau^{-\mathsf T}.
$
If clipping is inactive almost surely throughout a neighbourhood of
\(x_\star\), this limit is
\(H^{-1}\Sigma_0H^{-1}\), the ordinary nonprivate covariance.
\end{proposition}

At a fixed clipping radius, privacy therefore contributes a covariance term
that vanishes continuously with the randomisation noise.  Combined with
Proposition~\ref{prop:wp-clipping-moment}, this also separates the effects of
privacy noise and clipping bias.

\begin{theorem}\label{thm:wp-triangular-clt}
Let $\theta_n^\circ$ denote the population target associated with the $n$th clipping and privacy regime, and set $u_n=\Log_{\theta_n^\circ}(\widehat\theta_n)$.  Suppose there are nonsingular matrices $A_n$, deterministic positive-definite matrices $B_n$, and a martingale-difference array $\{\xi_{n,i},\mathcal F_{n,i}:1\leq i\leq n\}$ such that
\[
 B_n^{-1/2}A_n\sqrt n\,u_n
 =-\frac1{\sqrt n}\sum_{i=1}^n B_n^{-1/2}\xi_{n,i}+o_{\mathbb P}(1).
\]
Assume, for every $\eta>0$,
\begin{align*}
 &\frac1n\sum_{i=1}^n
 \mathbb E\!\left[B_n^{-1/2}\xi_{n,i}\xi_{n,i}^{\mathsf T}B_n^{-1/2}
 \mid\mathcal F_{n,i-1}\right]\ \xrightarrow{\mathbb P}\ I,\\
 &\frac1n\sum_{i=1}^n
 \mathbb E\!\left[\|B_n^{-1/2}\xi_{n,i}\|^2
 \mathbf 1\{\|B_n^{-1/2}\xi_{n,i}\|>\eta\sqrt n\}
 \mid\mathcal F_{n,i-1}\right]\ \xrightarrow{\mathbb P}\ 0.
\end{align*}
Then
\[
 B_n^{-1/2}A_n\sqrt n\,u_n\rightsquigarrow N(0,I).
\]
If positive-definite $\widehat B_n$ and $\widehat A_n$ additionally satisfy
\[
 \|\widehat B_n^{-1/2}B_n^{1/2}-I\|_{\mathrm{op}}=o_{\mathbb P}(1),\qquad
 \|\widehat B_n^{-1/2}\left(\widehat A_n-A_n\right)A_n^{-1}B_n^{1/2}\|_{\mathrm{op}}
 =o_{\mathbb P}(1),
\]
then
\[
 \widehat B_n^{-1/2}\widehat A_n\sqrt n\,u_n\rightsquigarrow N(0,I),
 \qquad
 n\,u_n^{\mathsf T}\widehat A_n^{\mathsf T}\widehat B_n^{-1}\widehat A_n u_n
 \rightsquigarrow\chi^2_d.
\]
\end{theorem}

Here $A_n$ and $B_n$ are the derivative and released-score covariance for a general estimating equation; for the gradient setting in the main text they correspond to $H_n$ and $\Sigma_{\mathrm{tot},n}$. The normalisation follows $B_n$ directly, so it may change with the clipping radius and privacy budget. A changing-regime implementation verifies the displayed linearization, Lindeberg condition, covariance convergence, and estimator rates.

\section{Simulation details and additional results}\label{app:sim-af}

This appendix records the six data-generating models used in
Section~\ref{sec:sim}.  Every reported confidence region uses the Hessian and
covariance fitted in
\eqref{eq:pair-regression}--\eqref{eq:wald-guard} from the released transcript.

\paragraph{Design A: directional location on \(\sph^2\).}
Let \(e_1=(1,0,0)^{\mathsf T}\), \(c=0.95\), and
\(s=(1-c^2)^{1/2}\).  The observation takes values
\[
 e_1,\quad(c,\pm s,0)^{\mathsf T},\quad(c,0,\pm s)^{\mathsf T}
\]
with probabilities \(0.4,0.15,0.15,0.15,0.15\).  The loss is
\(-5\mu^{\mathsf T}Y\), the target is \(e_1\), and the score bound is
\(G=5\).  Error is geodesic distance on the sphere.

\paragraph{Design B: bounded streaming PCA on \(\sph^3\).}
Generate
\[
 Y=\diag\left(\sqrt3,\sqrt{1.5},\sqrt{0.8},\sqrt{0.3}\right)\,U,
\]
where the coordinates of \(U\) are independent Rademacher variables.  The
target is the leading eigenvector \(e_1\).  For the Rayleigh loss,
\(\norm{\nabla\ell(u,Y)}\le\norm{Y}^2/2=2.8\), which is used for privacy
calibration.  Error is the sign-aligned principal angle.

\paragraph{Design C: log-Euclidean mean on \(\SPD_2\).}
Represent \(\log Y\) in the orthonormal basis
\[
 E_1=\begin{pmatrix}1&0\\0&0\end{pmatrix},\quad
 E_2=2^{-1/2}\begin{pmatrix}0&1\\1&0\end{pmatrix},\quad
 E_3=\begin{pmatrix}0&0\\0&1\end{pmatrix}.
\]
With \(s_\star=(0.2,0.1,-0.3)^{\mathsf T}\), the log coordinate takes values
\[
 s_\star,\quad s_\star\pm0.31e_1,\quad
 s_\star\pm0.38e_2,\quad s_\star\pm0.29e_3
\]
with probabilities \(0.4,0.1,\ldots,0.1\).  The implemented score is
\(s-\log Y\), so its Hessian is \(I_3\), although the formal procedure still
estimates it from the private message pairs.  Error is Euclidean distance in
log coordinates.

\paragraph{Design D: intrinsic Fr\'echet mean on \(\HH^2\).}
In the hyperboloid model, the support is
\[
 \Exp_{\mu_\star}(\pm0.60e_1),\qquad
 \Exp_{\mu_\star}(\pm0.35e_2),
\]
with probability \(1/4\) at each point.  The working ball has radius \(0.80\).
The distance between an admissible query and a support point is at most
\(0.80+0.60=1.40\), the score bound used for privacy calibration.  Error is
hyperbolic distance.

\paragraph{Design E: pseudo-Huber center on \(\HH^2\).}
The loss is
\[
 \ell(x,y)=0.45^2\left\{\sqrt{1+d(x,y)^2/0.45^2}-1\right\}.
\]
The support points are
\(\Exp_{\mu_\star}(\pm0.30e_1)\), with total probability \(0.7\), and
\(\Exp_{\mu_\star}(\pm1.00e_2)\), with total probability \(0.3\).
The radial derivative is bounded by \(G=0.45\).  Error is hyperbolic distance.

\paragraph{Design F: affine score in local SPD coordinates.}
This experiment isolates the three-dimensional local score associated with
the covariance quasi-likelihood.  At \(w\in\R^3\),
\[
 g(w,\xi)=\frac12w+\frac12
 \begin{pmatrix}1-\xi_1^2\\-\sqrt2\,\xi_1\xi_2\\1-\xi_2^2\end{pmatrix},
\]
where the coordinates of \(\xi\) are independent and take \(0\) with
probability \(1/2\) and \(\pm\sqrt2\) with probability \(1/4\) each.  The
target is \(w_\star=0\), \(H=I_3/2\), and the public coordinate ball gives
\(G=1+\sqrt{2.5}\).  The experiment uses ordinary addition in this declared
local chart; error is \(\norm{\bar w-w_\star}\).

\subsection{Implementation and tuning parameters}

For every design, the pair directions cycle through
\((e_1,-e_1,\ldots,e_d,-e_d)\).  The pair gain is
\[
 \gamma_k=2C(2k+b_\eps)^{-0.60},\qquad
 b_\eps=\max\!\left[0,
 \left\{2C\sigma\sqrt{d/2}/\delta_0\right\}^{1/0.60}-2\right],
\]
where \(C\) and the user-step cap \(\delta_0\) are given below.  For the
nonlinear designs, if \(c(k)\) denotes the completed direction-cycle index,
the radius is
\[
 h_k=s\min\{0.40,r(2c(k))^{-0.27}\}.
\]
Designs C and F instead use fixed radii \(0.14\) and \(0.30\).

\begin{table}[!htbp]
\centering
\caption{Point and query specifications.  ``Cap'' gives a spherical
or hyperbolic ball; ``box'' and ``ball'' are in the displayed local
coordinates.}
\label{tab:frozen-point-radius}
\begin{tabular}{@{}cllllrr@{}}
\toprule
Design & \(d\) & Initial point & Working region & \(s\) & \(r\) & \(C\)\\
\midrule
A & 2 & \(\operatorname{normalize}(.96,.23,-.16)\) & spherical cap \(x_1\ge .75\) & .60 & 3.5 & .8\\
B & 3 & \(\operatorname{normalize}(.95,.18,-.12,.20)\) & spherical cap \(x_1\ge .75\) & .60 & 3.5 & 1.0\\
C & 3 & \((.32,0,-.22)\) & box \(s_\star\pm .50\) & .35 & -- & .8\\
D & 2 & \(\Exp_0(.18,-.14)\) & hyperbolic cap, radius .80 & .45 & 3.5 & .8\\
E & 2 & \(\Exp_0(.18,-.12)\) & hyperbolic cap, radius 1.25 & .60 & 4.5 & .8\\
F & 3 & \((.20,-.10,.15)\) & Euclidean ball, radius 2 & .75 & -- & 1.0\\
\bottomrule
\end{tabular}
\end{table}

The state and query regions used in the code give an explicit feasibility
check for Assumption~\ref{ass:local}.  In A and B, the state is projected to
the cap \(x_1\ge .75\), whose angular radius is
\(\arccos(.75)=0.723\).  The nominal radius never exceeds .24, and the
normalised query \(\operatorname{normalize}(x\pm hv)\) is at geodesic distance
\(\arctan(h)\le0.236\) from \(x\); hence both queries lie in the public cap of
radius 1.00 about \(e_1\).  In C, the fixed radius is .14 and the update is
projected coordinatewise to \(s_\star\pm(.50-.14)\), so the two queries lie in
the box \(s_\star\pm.50\).  In D and E, respectively, \(h_k\le.18\) and
\(h_k\le.24\); the update is projected to the hyperbolic ball of radius
\(R-h_k\), with \(R=.80\) and \(R=1.25\), before the next pair is queried.
The triangle inequality therefore places both queries in the radius-\(R\)
ball.  These state balls are contained in query balls of radii 1.00 and 1.50.
Finally, F uses \(h=.30\), projects the state to the Euclidean ball of radius
1.70, and queries within the radius-2 ball.  All these projections are radial
or coordinatewise about public working-region centres.  In the controlled
designs those centres coincide with the population targets, which verifies
target-quasi-nonexpansiveness; neither the Hessian nor the score covariance is
supplied to the procedure.  The same query regions are used when evaluating the score bounds
that calibrate the Gaussian mechanisms.  Since \(h_k\) is fixed in C and F
and is deterministic and constant within each completed cycle in A, B, D,
and E, these feasibility projections preserve the exact balanced-cycle Gram
identities. These safeguards fall under the extension in Section~\ref{app:alternative-safeguards}.

The user-step caps are \(0.85\) in A and D, \(0.75\) in C, and
\(0.85\) in F when \(\eps\in\{\infty,8\}\), changing in F to
\(0.10,0.20,0.40\) at \(\eps=4,2,1\).  In B the cap is \(1.60\) at
\(\eps=8\) and \(1.35\) at the other private levels, with \(1.35\) in the
nonprivate experiment.  In E it is \(1.00,1.00,2.00,1.50,2.00\) at
\(\eps=\infty,8,4,2,1\), respectively.

For a general transcript-based Hessian floor, strict positivity can
be enforced by replacing its raw value with
\(\min\left\{L_H/2,\max\left(\varepsilon_n,\widehat\kappa_n^{\rm raw}\right)\right\}\),
where \(\varepsilon_n>0\) decreases to zero.
The nuisance window starts at the first full cycle after
\(\lceil K^{1/3}\rceil\) pairs and ends at the final completed cycle.  With
\(Q_n\) from \eqref{eq:pair-regression}, the ridge is
\(0.1\{\operatorname{tr}(Q_n)/d\}n^{-1/2}\).  The Hessian eigenvalue floor is
\[
\min\!\left\{\frac{L_H}{2},\,
0.5\max\left\{1,\operatorname{mean}\left|\lambda\left(\widehat H_n^{\rm raw}\right)\right|\right\}n^{-1/4}\right\},
\]
where the public upper bounds \(L_H\) for A--F are \(6.25,5.6,1.25,
1.4/\tanh(1.4),1,\) and \(0.6\).  The private covariance floor is the known
Gaussian noise variance; the nonprivate floor is
\[
0.1\max\left\{1,\operatorname{mean}\left|\lambda\left(\widehat\Sigma_n^{\rm raw}\right)\right|\right\}n^{-1/3}.
\]

The finite-sample adjustment in \eqref{eq:wald-guard} uses \(c=0.35\)
in designs A--D and F, and \(c=0.50\) in E.  The multiplier
\(a_{\rm g}\) is zero in A and B, \(0.20\) in C, and \(0.40\) in F.
In D it is \(0.50\) under finite privacy and zero in the nonprivate case;
in E the corresponding values are \(1.50\) and \(2.50\).
These settings were fixed using a separate calibration experiment before
the 500-repetition experiments.

\subsection{Alternative covariance estimations}

Section~\ref{sec:sim-comparators} of the main paper gives the two algorithms,
their frozen numerical tuning, and their relation to the reported figures.
For the validity result below, let \(w_i\) be the coordinate displacement of
query \(i\) from the transported trajectory anchor, let \(\bar w_n\) be its
average over the regression window \(\mathcal J_n\), and put
\[
 Q_n^{\rm tr}=\sum_{i\in\mathcal J_n}
 (w_i-\bar w_n)(w_i-\bar w_n)^{\mathsf T}.
\]
The fitted regression contains an intercept, so its slope uses only the
centred regressors in \(Q_n^{\rm tr}\).

\begin{theorem}
\label{thm:comparator-validity}
Suppose the ordinary one-query averaged recursion satisfies the corresponding
local stability, smoothness, moment, and central limit conditions, with
limiting Hessian \(H\) and released-score covariance
\(\Sigma_{\mathrm{tot}}\).  For either Wald conclusion below, suppose in
addition that \(\Sigma_{\mathrm{tot}}\succ0\).

\begin{enumerate}
\item For same-trajectory regression, freeze the anchor before a
window chosen by the public clock, and let each query be predictable
before its release.  Suppose the anchor converges to \(x_\star\) and
\(\max_{i\in\mathcal J_n}\norm{w_i}\to_P0\),
\[
 \max_{i\in\mathcal J_n}\norm{w_i-\bar w_n}\xrightarrow[]{P}0,
 \qquad \lambda_{\min}(Q_n^{\rm tr})\xrightarrow[]{P}\infty,
 \qquad
 \frac{\lambda_{\max}(Q_n^{\rm tr})}
      {\lambda_{\min}(Q_n^{\rm tr})}=O_P(1),
\]
\(\sum_{i\in\mathcal J_n}\norm{w_i}^2=O_P\{\tr(Q_n^{\rm tr})\}\), and the
ridge is \(o_P\{\lambda_{\min}(Q_n^{\rm tr})\}\).  Then the fitted
Hessian and residual covariance are consistent for \(H\) and
\(\Sigma_{\mathrm{tot}}\).  The corresponding Wald statistic, scaled by the
number of users in the point recursion, converges to \(\chi_d^2\).

\item For half-sample public probes, let \(m_n\) and \(r_n=n-m_n\) be the
point and probe sample sizes, with \(m_n/n\to\rho\in(0,1)\).  Suppose the
probe directions form completed tight-frame cycles, their common radius
\(h_n\) satisfies \(h_n\to0\) and \(r_nh_n^2\to\infty\), and the ridge is
\(o(r_nh_n^2)\).  Then the probe estimates of \(H\) and
\(\Sigma_{\mathrm{tot}}\) are consistent, and the Wald statistic scaled by
\(m_n\) converges to \(\chi_d^2\).
\end{enumerate}
\end{theorem}

\subsection{Additional simulation results}

Only estimation error and feasible coverage are reported.

\begin{table}[!htbp]
\centering
\caption{Terminal mean error and empirical coverage of nominal 95\% feasible
Wald regions at \(n=32000\).  Each entry is based on 500 repetitions.}
\label{tab:sim-terminal-af}
\begin{tabular}{ccrr@{\qquad}ccrr}
\toprule
Design & \(\eps\) & Error & Coverage &
Design & \(\eps\) & Error & Coverage\\
\midrule
A & \(\infty\) & 0.0012 & 0.962 & D & \(\infty\) & 0.0023 & 0.954\\
A & 8            & 0.0099 & 0.950 & D & 8            & 0.0122 & 0.956\\
A & 4            & 0.0175 & 0.964 & D & 4            & 0.0231 & 0.952\\
A & 2            & 0.0236 & 0.990 & D & 2            & 0.0394 & 0.954\\
A & 1            & 0.0213 & 1.000 & D & 1            & 0.0464 & 1.000\\
\addlinespace
B & \(\infty\) & 0.0082 & 0.944 & E & \(\infty\) & 0.0025 & 0.944\\
B & 8            & 0.0196 & 0.948 & E & 8            & 0.0079 & 0.936\\
B & 4            & 0.0335 & 0.950 & E & 4            & 0.0143 & 0.942\\
B & 2            & 0.0396 & 0.996 & E & 2            & 0.0273 & 0.972\\
B & 1            & 0.0303 & 1.000 & E & 1            & 0.0539 & 0.962\\
\addlinespace
C & \(\infty\) & 0.0013 & 0.950 & F & \(\infty\) & 0.0107 & 0.950\\
C & 8            & 0.0135 & 0.936 & F & 8            & 0.0632 & 0.948\\
C & 4            & 0.0229 & 0.962 & F & 4            & 0.1109 & 0.968\\
C & 2            & 0.0348 & 0.994 & F & 2            & 0.1764 & 0.994\\
C & 1            & 0.0322 & 1.000 & F & 1            & 0.1632 & 1.000\\
\bottomrule
\end{tabular}
\end{table}

\section{Technical analysis of data privatisation}\label{app:proof-point}

Suppose the raw observation is privatised through a Markov kernel \(Q\),
producing \(Z\sim Q(\cdot\mid Y)\), and the analyst then applies the original
loss to \(Z\).  Write \(F_Q(x)=\E\{\ell(x,Z)\}\).

\begin{theorem}\label{thm:targetshift}
Let \(U\) be a normal convex neighbourhood of \(x_\star\). Assume
\(x\mapsto F_Q(x)\) is \(C^2\) on \(U\) and has an interior local minimiser
\(x_Q\in U\). Define
\[
        \bar F_Q(v)=F_Q(\Exp_{x_\star}v),\qquad
        \Delta_Q=\Log_{x_\star}(x_Q).
\]
Assume that
\begin{equation}\label{eq:AQ-def}
        A_Q=\int_0^1 \nabla^2\bar F_Q(t\Delta_Q)\,dt
\end{equation}
is nonsingular. Then
\begin{equation}\label{eq:shift-identity}
        \Delta_Q=-A_Q^{-1}\nabla\bar F_Q(0),
\end{equation}
and, for this local minimiser,
\begin{equation}\label{eq:target-coincidence}
        x_Q=x_\star
        \quad\Longleftrightarrow\quad
        \nabla F_Q(x_\star)=0.
\end{equation}
Moreover, consider a sequence of privatisation kernels for which
\(\Delta_Q\to0\). Suppose that \(F\) is \(C^2\) near \(x_\star\),
\(H=\nabla^2F(x_\star)\) is nonsingular,
\(\nabla^2\bar F_Q(0)\to H\), and the Hessians are locally equicontinuous at
zero: for some \(r>0\) and deterministic modulus \(\omega(t)\downarrow0\),
\[
 \norm{\nabla^2\bar F_Q(u)-\nabla^2\bar F_Q(v)}_{\op}
 \le \omega\left(\norm{u-v}\right)
\]
whenever \(\norm u,\norm v\le r\). Then
\begin{equation}\label{eq:first-order-shift}
        \Log_{x_\star}(x_Q)
        =-H^{-1}\nabla F_Q(x_\star)
          +o\left\{\norm{\nabla F_Q(x_\star)}\right\}.
\end{equation}
\end{theorem}

The next result describes the leading bias produced by a centred,
small-noise release. Let \((\mathcal W,\norm{\cdot}_{\mathcal W})\) be a
finite-dimensional normed vector space. For a conditional perturbation
\(W_\rho\), let \(T:\cY\times\mathcal W\to\cY\) satisfy \(T(y,0)=y\), set
\[
        Z_\rho=T(Y,W_\rho),\qquad
        \rho^2=\E\norm{W_\rho}_{\mathcal W}^2,
\]
and write \(F_{Q_\rho}\) for the corresponding population objective.

\begin{theorem}
\label{thm:smallnoise}
Work under the local smoothness assumptions of
Theorem~\ref{thm:targetshift}. Assume
\(\E(W_\rho\mid Y)=0\) and \(\rho^2\to0\), and define
\[
        h_Y(w)=\nabla_x\ell\{x_\star,T(Y,w)\}\in T_{x_\star}M.
\]
Assume \(h_Y(0)\) and \(h_Y(W_\rho)\) are integrable and differentiation
under the expectation is valid:
\begin{equation}\label{eq:smallnoise-diff-expectation}
        \nabla F(x_\star)=\E\{h_Y(0)\},\qquad
        \nabla F_{Q_\rho}(x_\star)=\E\{h_Y(W_\rho)\}.
\end{equation}
Suppose \(h_Y\) is twice Fr\'echet differentiable at zero almost surely and
\begin{equation}\label{eq:smallnoise-ui}
 \E\norm{h_Y(W_\rho)-h_Y(0)-Dh_Y(0)[W_\rho]
 -\tfrac12D^2h_Y(0)[W_\rho,W_\rho]}=o(\rho^2),
\end{equation}
with
\begin{equation}\label{eq:smallnoise-second-bound}
 \E\!\left\{\norm{D^2h_Y(0)}_{\op}
             \norm{W_\rho}_{\mathcal W}^2\right\}=O(\rho^2).
\end{equation}
Then
\begin{equation}\label{eq:gradient-bias}
 \nabla F_{Q_\rho}(x_\star)
 =\frac12\E\{D^2h_Y(0)[W_\rho,W_\rho]\}+o(\rho^2).
\end{equation}
If the expansion \eqref{eq:first-order-shift} also applies, then
\begin{equation}\label{eq:smallnoise-shift}
 \Log_{x_\star}(x_{Q_\rho})
 =-\frac12H^{-1}\E\{D^2h_Y(0)[W_\rho,W_\rho]\}+o(\rho^2).
\end{equation}
\end{theorem}

\section{Auxiliary results for SPR inference}\label{app:auxiliary-results}

All vectors in this section are expressed in the smooth public frame.  The
limiting frame is \(B_{x_\star}\); smooth changes between nearby frames are
orthogonal up to \(o_P(1)\).  Proofs not included here are given in the Supplementary Material.

\begin{lemma}\label{lem:primitive-expansions}
Under Assumptions~\ref{ass:local} and \ref{ass:pair-smooth},
\(H\succeq\mu I_d\).  On the compact query paths, \(D\Psi_x(w)\) is
uniformly Lipschitz in \(w\), and

\[
 D\Psi_x(0)=B_x^{-1}\nabla^2F(x)B_x.
\]

Let \(\phi(x)=R_{x_\star}^{-1}(x)\) in the limiting frame and
\(J(x)=D_0[u\mapsto\phi\{R_x(B_xu)\}]\).  Near \(x_\star\),

\[
 J(x_\star)=I_d,\qquad
 J(x)B_x^{-1}\nabla F(x)=H\phi(x)+O\{\norm{\phi(x)}^2\}.
\]

\end{lemma}

\begin{lemma}\label{lem:local-projection}
Let \(C\) be a compact geodesically convex subset of the normal
convex neighbourhood \(U\).  Its metric projection \(P_C\) is
single-valued and continuous on a sufficiently small tubular
neighbourhood of \(C\).  There are uniform finite constants \(A,C_R\)
such that, for \(t,x\in C\) and every sufficiently small \(u\in T_xM\),

\[
 d\{P_C(y),t\}^2\le d(y,t)^2+A d\{y,P_C(y)\}^2,
 \qquad y=R_x(u),\qquad d(x,y)\le C_R\norm u.
\]

Consequently, for \(V_t(x)=d(x,t)^2\),

\[
 V_t\{P_C R_x(u)\}
 \le V_t(x)+D V_t(x)[u]+A'\norm u^2.
\]

\end{lemma}

\begin{corollary}\label{cor:pair-covariance-limits}
Under the assumptions of Theorem~\ref{thm:primitive-pair-stability},
\begin{equation}\label{eq:proof-pair-covariance-limits}
 \Cov(\xi_k\mid\mathcal G_{k-1})\to_P\frac12\Sigma_{\mathrm{tot}},
 \qquad
 \Cov(\zeta_k\mid\mathcal G_{k-1})\to_P\frac12\Sigma_{\mathrm{tot}}.
\end{equation}
\end{corollary}

\begin{lemma}\label{lem:step-cap}
Let \(W_k\) be measurable after pair \(k\) and satisfy
\(\E\left(\|W_k\|^4\,\middle|\,\mathcal G_{k-1}\right)\le C\).  For
\(\widetilde W_k=\mathsf C_{\delta_0/\gamma_k}(W_k)\), uniformly in \(k\),
\begin{align}
 \Pr\left(\widetilde W_k\ne W_k\,\middle|\,\mathcal G_{k-1}\right)&\le C\gamma_k^4,\notag\\
 \left\|\E\left(\widetilde W_k-W_k\,\middle|\,\mathcal G_{k-1}\right)\right\|
 &\le C\gamma_k^3,\label{eq:cap-mean}\\
 \left\|\E\left(\widetilde W_k\widetilde W_k^{\mathsf T}
       -W_kW_k^{\mathsf T}\,\middle|\,\mathcal G_{k-1}\right)\right\|_{\op}
 &\le C\gamma_k^2.\label{eq:cap-covariance}
\end{align}
Moreover \(\gamma_k\|\widetilde W_k\|\le\delta_0\) deterministically.
\end{lemma}

\begin{lemma}
\label{lem:target-frame-expansion}
Let \(\phi(x)=R_{x_\star}^{-1}(x)\), expressed in the limiting frame, and set
\[
 J(x)=D_0\bigl[u\mapsto \phi\{R_x(B_xu)\}\bigr].
\]
Let \(m_K=\lfloor K^\kappa\rfloor\), \(0<\kappa<1/2\).  On the event in
\eqref{eq:tail-safeguard-inactive}, every update after \(m_K\) admits the
expansion
\begin{equation}\label{eq:target-frame-recursion}
 \Delta_k=\Delta_{k-1}-\gamma_k
 \{H\Delta_{k-1}+\xi_k^\star+c_k\}+\rho_k,
 \qquad \xi_k^\star=J(x_{k-1})\widetilde\xi_k,
\end{equation}
where
\(\widetilde\xi_k=\widetilde U_k-
\E\left(\widetilde U_k\,\middle|\,\mathcal G_{k-1}\right)\) and \(\xi_k^\star\) is a martingale
difference.  Uniformly for \(k>m_K\),
\begin{align}
 \norm{c_k}&\le C\left\{\norm{\Delta_{k-1}}^2+h_k^2+\gamma_k^3\right\},
 \label{eq:target-frame-drift-bound}\\
 \E\left(\norm{\rho_k}\,\middle|\,\mathcal G_{k-1}\right)&\le C\gamma_k^2.
 \label{eq:target-frame-step-bound}
\end{align}
Moreover,
\begin{equation}\label{eq:intrinsic-average-equivalence}
 \norm{R_{x_\star}^{-1}(\bar x_K)
       -K^{-1}\sum_{k=1}^K\Delta_k}
 \le \frac{C}{K}\sum_{k=1}^K\norm{\Delta_k}^2+o_P(K^{-1/2}).
\end{equation}
Under Assumption~\ref{ass:pair-design}, consequently,
\begin{equation}\label{eq:target-frame-small-terms}
 \frac1{\sqrt K}\sum_{k=m_K+1}^K\norm{c_k}=o_P(1),\qquad
 \frac1{\sqrt K}\sum_{k=m_K+1}^K\frac{\norm{\rho_k}}{\gamma_k}=o_P(1),
\end{equation}
and the right side of \eqref{eq:intrinsic-average-equivalence} is
\(o_P(K^{-1/2})\).
\end{lemma}

\begin{lemma}
\label{lem:perturbed-pr}
Suppose a bounded tangent recursion has the form, on an event with
probability tending to one, for \(m_K<k\le K\), where
\(m_K=o(\sqrt K)\),

\begin{equation}\label{eq:proof-local-recursion}
 \Delta_k=\Delta_{k-1}-\gamma_k
       \{H\Delta_{k-1}+\xi_k+b_k\}+\rho_k,
\end{equation}

where \(H\) is invertible, \(\xi_k\) is a martingale difference with a
uniformly bounded fourth moment,
\(\gamma_k=\gamma_0(k+k_0)^{-\alpha}\), \(1/2<\alpha<1\), and the stability
bounds in \eqref{eq:pair-stability} hold with one of the two radius regimes
in Assumption~\ref{ass:pair-design}.  Assume

\[
 \frac1{\sqrt K}\sum_{k=m_K+1}^K\norm{b_k}\to0,
 \qquad
 \frac1{\sqrt K}\sum_{k=m_K+1}^K\frac{\norm{\rho_k}}{\gamma_k}
 \xrightarrow[]{P}0.
\]

Then the intrinsic Polyak--Ruppert average satisfies

\begin{equation}\label{eq:proof-pr-linear}
 \sqrt K\,R_{x_\star}^{-1}(\bar x_K)
 =-H^{-1}\frac1{\sqrt K}\sum_{k=m_K+1}^K\xi_k+o_P(1).
\end{equation}
\end{lemma}

\subsection{Alternative public safeguards}\label{app:alternative-safeguards}

The fixed metric projection may be replaced by public predictable
maps \(\Pi_k\) into \(K_{\rm s}\).  For this extension, require that the
maps are measurable, that every capped trial lies in their geometric
domain, and that, uniformly in the row and update index,

\[
 d\{\Pi_k(y),x_\star\}\le d(y,x_\star),\qquad
 \Pi_k(y)=y\quad\text{for }y\in K_0,
\]

where the fixed set \(K_0\subset K_{\rm s}\) contains a neighbourhood
of \(x_\star\).  The safeguarded states must retain the prescribed
query paths in \(K_{\rm q}\); the maps do not change the public radii,
directions, or complete-cycle regression window.  Under these conditions,
the conclusions for the averaged estimator, SPR, and Wald inference
remain valid.  The fixed-radius conclusions retain their separate
bias and regression conditions.

Indeed, target-quasi-nonexpansiveness gives the Lyapunov inequality
in Lemma~\ref{lem:local-projection} directly, so the second- and
fourth-moment recursions are unchanged.  Choose \(\rho>0\) with
\(B_{3\rho}(x_\star)\subset K_0\).  The tail localization bound and
\(d(x,R_xu)\le C_R\norm u\), followed by the fourth-moment union bound,
make every tail trial lie in \(K_0\) with probability tending to one.
All subsequent local expansions and regression identities therefore apply.
This extension includes the radius-dependent inner projections used in
the controlled simulations when their ranges contain one fixed target
neighbourhood.

\section{Construction of the diffusion-tensor illustration}
\label{sec:dti-construction}

The image and neighbouring-voxel tensors come from the Stanford HARDI data
(Rokem et al., 2013; PDDL~1.0).  Each fitted diffusion tensor is a
\(3\times3\) positive-definite matrix.  Its physical ellipsoid has principal
axes along the eigenvectors and half-axis lengths proportional to the square
roots of the eigenvalues.  The teal glyph marks the selected tensor, which
is also shown enlarged.  The neighbouring glyphs are voxels from the same
participant.

The cone uses separate synthetic \(2\times2\) matrices.  For
\[
 X=\begin{pmatrix}a&b\\b&c\end{pmatrix},\qquad
 x=\frac{a-c}{\sqrt2},\quad y=\sqrt2b,\quad z=\frac{a+c}{\sqrt2},
\]
these orthonormal coordinates identify \(\mathrm{SPD}(2)\) with the open
cone \(z>\sqrt{x^2+y^2}\).  Its lateral surface consists of singular
positive-semidefinite matrices; the upper rim is a display truncation.
The orange centre is the affine-invariant mean of the synthetic tensors.
The surrounding shell is formed by mapping a tangent ellipsoid through
\[
 \Exp_X(V)=X^{1/2}\exp\{X^{-1/2}VX^{-1/2}\}X^{1/2}.
\]
The shell is cut away to show its centre, and the cone is scaled for display.
This is a schematic uncertainty region without a calibrated coverage level.
The three-dimensional cone illustrates positive-definite geometry; it is not
a projection of the six-dimensional \(\mathrm{SPD}(3)\) brain tensors.

The LDP arrow represents a population protocol in which each independent
participant retains one tensor at a common public registered location and
releases a randomised tangent score \(\widetilde g=g+\xi\).  The synthetic
cone illustrates estimation and inference from such scores; it is not a
private population analysis of the displayed MRI data.


\begin{thebibliography}{99}

\bibitem[Absil et~al.(2008)Absil, Mahony, and Sepulchre]{Absil2008}
P.-A. Absil, R. Mahony, and R. Sepulchre.
\newblock \emph{Optimization Algorithms on Matrix Manifolds}.
\newblock Princeton University Press, 2008.

\bibitem[Afsari(2011)]{Afsari2011}
B. Afsari.
\newblock Riemannian $L^p$ center of mass: Existence, uniqueness, and convexity.
\newblock \emph{Proceedings of the American Mathematical Society}, 139(2):655--673, 2011.

\bibitem[Arsigny et~al.(2006)Arsigny, Fillard, Pennec, and Ayache]{Arsigny2006}
V. Arsigny, P. Fillard, X. Pennec, and N. Ayache.
\newblock Log-Euclidean metrics for fast and simple calculus on diffusion tensors.
\newblock \emph{Magnetic Resonance in Medicine}, 56(2):411--421, 2006.

\bibitem[Arsigny et~al.(2007)Arsigny, Fillard, Pennec, and Ayache]{Arsigny2007}
V. Arsigny, P. Fillard, X. Pennec, and N. Ayache.
\newblock Geometric means in a novel vector space structure on symmetric positive-definite matrices.
\newblock \emph{SIAM Journal on Matrix Analysis and Applications}, 29(1):328--347, 2007.

\bibitem[Ashwell et~al.(2012)Ashwell, Gunn, and Gibson]{Ashwell2012}
M. Ashwell, P. Gunn, and S. Gibson.
\newblock Waist-to-height ratio is a better screening tool than waist circumference and {BMI} for adult cardiometabolic risk factors: Systematic review and meta-analysis.
\newblock \emph{Obesity Reviews}, 13(3):275--286, 2012.
\newblock doi:10.1111/j.1467-789X.2011.00952.x.

\bibitem[Balle and Wang(2018)]{BalleWang2018}
B. Balle and Y.-X. Wang.
\newblock Improving the Gaussian mechanism for differential privacy: Analytical calibration and optimal denoising.
\newblock In \emph{Proceedings of the 35th International Conference on Machine Learning}, volume 80 of \emph{Proceedings of Machine Learning Research}, pages 394--403, 2018.

\bibitem[Banerjee et~al.(2005)Banerjee, Dhillon, Ghosh, and Sra]{Banerjee2005}
A. Banerjee, I. S. Dhillon, J. Ghosh, and S. Sra.
\newblock Clustering on the unit hypersphere using von {M}ises--{F}isher distributions.
\newblock \emph{Journal of Machine Learning Research}, 6:1345--1382, 2005.

\bibitem[Bhatia(2009)]{Bhatia2009}
R. Bhatia.
\newblock \emph{Positive Definite Matrices}.
\newblock Princeton University Press, 2009.

\bibitem[Bhattacharya and Patrangenaru(2003)]{BhattacharyaPatrangenaru2003}
R. Bhattacharya and V. Patrangenaru.
\newblock Large sample theory of intrinsic and extrinsic sample means on manifolds. I.
\newblock \emph{The Annals of Statistics}, 31(1):1--29, 2003.

\bibitem[Bhattacharya and Patrangenaru(2005)]{BhattacharyaPatrangenaru2005}
R. Bhattacharya and V. Patrangenaru.
\newblock Large sample theory of intrinsic and extrinsic sample means on manifolds. II.
\newblock \emph{The Annals of Statistics}, 33(3):1225--1259, 2005.

\bibitem[Bonnabel(2013)]{Bonnabel2013}
S. Bonnabel.
\newblock Stochastic gradient descent on Riemannian manifolds.
\newblock \emph{IEEE Transactions on Automatic Control}, 58(9):2217--2229, 2013.

\bibitem[Boumal(2023)]{Boumal2023}
N. Boumal.
\newblock \emph{An Introduction to Optimization on Smooth Manifolds}.
\newblock Cambridge University Press, 2023.

\bibitem[Cai et~al.(2025)Cai, Hu, Sun, and Wu]{CaiHuSunWu2025}
L. Cai, Q. Hu, J. Sun, and S. Wu.
\newblock Time-uniform and asymptotic confidence sequence of quantile under local differential privacy.
\newblock In \emph{Advances in Neural Information Processing Systems}, volume 38, pages 114488--114520, 2025.

\bibitem[Cai et~al.(2026)Cai, Hu, and Wu]{CaiHuWu2026Federated}
L. Cai, Q. Hu, and S. Wu.
\newblock Federated learning of quantile inference under local differential privacy.
\newblock In \emph{International Conference on Learning Representations}, 2026.

\bibitem[Chang et~al.(2026)Chang, Jiang, Mostajeran, and Hu]{ChangEtAl2026RDP}
X. Chang, Y. Jiang, C. Mostajeran, and Q. Hu.
\newblock Geometric R\'enyi differential privacy: Ricci curvature characterized by heat diffusion mechanisms.
\newblock arXiv:2604.20761, 2026.

\bibitem[Chen et~al.(2020)Chen, Clark, Riddles, Mohadjer, and Fakhouri]{CDCDesign2020}
T.-C. Chen, J. Clark, M. K. Riddles, L. K. Mohadjer, and T. H. I. Fakhouri.
\newblock National Health and Nutrition Examination Survey, 2015--2018: Sample design and estimation procedures.
\newblock \emph{Vital and Health Statistics}, Series 2, No. 184, 2020.

\bibitem[Duchi et~al.(2018)Duchi, Jordan, and Wainwright]{DuchiJordanWainwright2018}
J. C. Duchi, M. I. Jordan, and M. J. Wainwright.
\newblock Minimax optimal procedures for locally private estimation.
\newblock \emph{Journal of the American Statistical Association}, 113(521):182--201, 2018.

\bibitem[Edelman et~al.(1998)Edelman, Arias, and Smith]{Edelman1998}
A. Edelman, T. A. Arias, and S. T. Smith.
\newblock The geometry of algorithms with orthogonality constraints.
\newblock \emph{SIAM Journal on Matrix Analysis and Applications}, 20(2):303--353, 1998.

\bibitem[Fr\'echet(1948)]{Frechet1948}
M. Fr\'echet.
\newblock Les \'el\'ements al\'eatoires de nature quelconque dans un espace distanci\'e.
\newblock \emph{Annales de l'Institut Henri Poincar\'e}, 10(4):215--310, 1948.

\bibitem[Han et~al.(2024)Han, Mishra, Jawanpuria, and Gao]{Han2024}
A. Han, B. Mishra, P. Jawanpuria, and J. Gao.
\newblock Differentially private Riemannian optimization.
\newblock \emph{Machine Learning}, 113(3):1133--1161, 2024.

\bibitem[Jiang et~al.(2023)Jiang, Chang, Liu, Ding, Kong, and Jiang]{Jiang2023}
Y. Jiang, X. Chang, Y. Liu, L. Ding, L. Kong, and B. Jiang.
\newblock Gaussian differential privacy on Riemannian manifolds.
\newblock In \emph{Advances in Neural Information Processing Systems}, volume 36, pages 14665--14684, 2023.

\bibitem[Jiang et~al.(2024)Jiang, Liu, Yan, Charest, Kong, and Jiang]{JiangMeasurement2024}
Y. Jiang, Y. Liu, X. Yan, A.-S. Charest, L. Kong, and B. Jiang.
\newblock Analysis of differentially private synthetic data: A measurement error approach.
\newblock \emph{Proceedings of the AAAI Conference on Artificial Intelligence}, 38(19):21206--21213, 2024.
\newblock doi:10.1609/aaai.v38i19.30114.

\bibitem[Jiang et~al.(2026)Jiang, Chang, and Hu]{JiangChangHu2026}
Y. Jiang, X. Chang, and Q. Hu.
\newblock Differentially private inference framework for Riemannian manifold data.
\newblock arXiv:2605.14762, 2026.

\bibitem[Jolliffe and Cadima(2016)]{JolliffeCadima2016}
I. T. Jolliffe and J. Cadima.
\newblock Principal component analysis: A review and recent developments.
\newblock \emph{Philosophical Transactions of the Royal Society A}, 374(2065):20150202, 2016.

\bibitem[Kairouz et~al.(2016)Kairouz, Oh, and Viswanath]{KairouzOhViswanath2016}
P. Kairouz, S. Oh, and P. Viswanath.
\newblock Extremal mechanisms for local differential privacy.
\newblock \emph{Journal of Machine Learning Research}, 17(17):1--51, 2016.

\bibitem[Liu et~al.(2023)Liu, Hu, Ding, and Kong]{Liu2023}
Y. Liu, Q. Hu, L. Ding, and L. Kong.
\newblock Online local differential private quantile inference via self-normalization.
\newblock In \emph{Proceedings of the 40th International Conference on Machine Learning}, volume 202 of \emph{Proceedings of Machine Learning Research}, pages 21698--21714, 2023.

\bibitem[Mardia and Jupp(2000)]{MardiaJupp2000}
K. V. Mardia and P. E. Jupp.
\newblock \emph{Directional Statistics}.
\newblock Wiley, 2000.

\bibitem[National Center for Health Statistics(2020)]{CDCBodyMeasures2020}
National Center for Health Statistics.
\newblock National Health and Nutrition Examination Survey 2017--2018: Body measures (BMX\_J).
\newblock Centers for Disease Control and Prevention, 2020.
\newblock \url{https://wwwn.cdc.gov/Nchs/Data/Nhanes/Public/2017/DataFiles/BMX_J.htm}.

\bibitem[Pelletier(1998)]{Pelletier1998}
M. Pelletier.
\newblock On the almost sure asymptotic behaviour of stochastic algorithms.
\newblock \emph{Stochastic Processes and their Applications}, 78(2):217--244, 1998.
\newblock doi:10.1016/S0304-4149(98)00029-5.

\bibitem[Pennec(2006)]{Pennec2006}
X. Pennec.
\newblock Intrinsic statistics on Riemannian manifolds: Basic tools for geometric measurements.
\newblock \emph{Journal of Mathematical Imaging and Vision}, 25(1):127--154, 2006.

\bibitem[Reimherr et~al.(2021)Reimherr, Bharath, and Soto]{Reimherr2021}
M. Reimherr, K. Bharath, and C. Soto.
\newblock Differential privacy over Riemannian manifolds.
\newblock In \emph{Advances in Neural Information Processing Systems}, volume 34, pages 12292--12303, 2021.

\bibitem[Ruppert(1985)]{Ruppert1985}
D. Ruppert.
\newblock A Newton--Raphson version of the multivariate Robbins--Monro procedure.
\newblock \emph{The Annals of Statistics}, 13(1):236--245, 1985.
\newblock doi:10.1214/aos/1176346589.

\bibitem[Soto et~al.(2022)Soto, Bharath, Reimherr, and Slavkovi\'c]{Soto2022}
C. Soto, K. Bharath, M. Reimherr, and A. Slavkovi\'c.
\newblock Shape and structure preserving differential privacy.
\newblock In \emph{Advances in Neural Information Processing Systems}, volume 35, pages 24693--24705, 2022.

\bibitem[Spall(2000)]{Spall2000}
J. C. Spall.
\newblock Adaptive stochastic approximation by the simultaneous perturbation method.
\newblock \emph{IEEE Transactions on Automatic Control}, 45(10):1839--1853, 2000.
\newblock doi:10.1109/TAC.2000.880982.

\bibitem[Tripuraneni et~al.(2018)Tripuraneni, Flammarion, Bach, and Jordan]{Tripuraneni2018}
N. Tripuraneni, N. Flammarion, F. Bach, and M. I. Jordan.
\newblock Averaging stochastic gradient descent on Riemannian manifolds.
\newblock In \emph{Proceedings of the 31st Conference on Learning Theory}, volume 75 of \emph{Proceedings of Machine Learning Research}, pages 650--687, 2018.

\bibitem[Utpala et~al.(2023)Utpala, Vepakomma, and Miolane]{Utpala2023}
S. Utpala, P. Vepakomma, and N. Miolane.
\newblock Differentially private Fr\'echet mean on the manifold of symmetric positive definite ({SPD}) matrices with log-Euclidean metric.
\newblock \emph{Transactions on Machine Learning Research}, 2023.

\bibitem[van der Vaart(1998)]{wpVanderVaart1998}
A. W. van der Vaart.
\newblock \emph{Asymptotic Statistics}.
\newblock Cambridge University Press, Cambridge, 1998.

\bibitem[Warner(1965)]{Warner1965}
S. L. Warner.
\newblock Randomized response: A survey technique for eliminating evasive answer bias.
\newblock \emph{Journal of the American Statistical Association}, 60(309):63--69, 1965.

\bibitem[Xie et~al.(2025)Xie, Shi, Jiang, Kong, and He]{XieEtAl2025}
J. Xie, E. Shi, B. Jiang, L. Kong, and X. He.
\newblock Online differentially private inference in stochastic gradient descent.
\newblock arXiv:2505.08227, 2025.

\end{thebibliography}
\end{document}